\documentclass[journal,comsoc]{IEEEtran}

\usepackage[T1]{fontenc}
\usepackage{epsfig}
\usepackage{amsmath,bm}
\usepackage{algorithm}
\usepackage{algorithmic}
\usepackage{subfigure}
\usepackage{epstopdf}
\usepackage{enumerate}
\usepackage{tabularx}
\usepackage{amsthm}
\usepackage{booktabs}
\usepackage{balance}

\newtheorem{definition}{Definition}

\ifCLASSINFOpdf
\else
\fi

\usepackage{amsmath}
\usepackage[cmintegrals]{newtxmath}

\begin{document}
\title{Anomaly Detection on Attributed Networks via Contrastive Self-Supervised Learning}

\author{Yixin Liu,~
        Zhao Li,~
        Shirui Pan, \IEEEmembership{Member,~IEEE},
        Chen Gong, \IEEEmembership{Member,~IEEE}, \\
        Chuan Zhou, 
        George Karypis, \IEEEmembership{Fellow,~IEEE}

\thanks{Manuscript received November 30, 2020; revised February 26, 2021; accepted March 19, 2021. This work was supported in part by the National Natural Science Foundation (NSF) of China under Grant 61973162 and Grant 61872360, in part by the Fundamental Research Funds for the Central Universities under Grant 30920032202, in part by the China Computer Federation (CCF)-Tencent Open Fund under Grant RAGR20200101, and in part by the Hong Kong Scholars Program under Grant XJ2019036. \textit{(Yixin Liu and Zhao Li contributed equally to this work.) (Corresponding author: Shirui Pan.)}}
 \thanks{Yixin Liu and Shirui Pan are with the Department of Data Science and AI, Faculty of Information Technology, Monash University, Clayton, VIC 3800, Australia (e-mail: yixin.liu@monash.edu; shirui.pan@monash.edu).}
 \thanks{Zhao Li is with the Alibaba Group, Hangzhou 310000, China (e-mail: lizhao.lz@alibaba-inc.com).}
 \thanks{Chen Gong is with the PCA Laboratory, Key Laboratory of Intelligent Perception and Systems for High-Dimensional Information of Ministry of Education, School of Computer Science and Engineering, Nanjing University of Science and Technology, Nanjing 210094, China, and also with the Department of Computing, The Hong Kong Polytechnic University, Hong Kong (e-mail: chen.gong@njust.edu.cn).}
 \thanks {Chuan Zhou is with the Academy of Mathematics and Systems Science, Chinese Academy of Sciences, Beijing 100093, China (e-mail: zhouchuan@amss.ac.cn).}
 \thanks{George Karypis is with the Department of Computer Science and Engineering, University of Minnesota, Minneapolis, MN 55455 USA (e-mail: karypis@umn.edu).}
\thanks{Color versions of one or more figures in this article are available at https://doi.org/10.1109/TNNLS.2021.3068344.}
\thanks{Digital Object Identifier 10.1109/TNNLS.2021.3068344}
}

\markboth{IEEE TRANSACTIONS ON NEURAL NETWORKS AND LEARNING SYSTEMS}%
{Shell \MakeLowercase{\textit{et al.}}: Bare Demo of IEEEtran.cls for IEEE Communications Society Journals}

\maketitle

\begin{abstract}
Anomaly detection on attributed networks attracts considerable research interests due to wide applications of attributed networks in modeling a wide range of complex systems. Recently, the deep learning-based anomaly detection methods have shown promising results over shallow approaches, especially on networks with high-dimensional attributes and complex structures. However, existing approaches, which employ graph autoencoder as their backbone, do not fully exploit the rich information of the network, resulting in suboptimal performance. Furthermore, these methods do not directly target anomaly detection in their learning objective and fail to scale to large networks due to the full graph training mechanism. To overcome these limitations, in this paper, we present a novel \underline{\textbf{Co}}ntrastive self-supervised \underline{\textbf{L}}earning framework for \underline{\textbf{A}}nomaly detection on attributed networks (\textbf{CoLA} for abbreviation). Our framework fully exploits the local information from network data by sampling a novel type of contrastive instance pair, which can capture the relationship between each node and its neighboring substructure in an unsupervised way. Meanwhile, a well-designed graph neural network-based contrastive learning model is proposed to learn informative embedding from high-dimensional attributes and local structure and measure the agreement of each instance pairs with its outputted scores. The multi-round predicted scores by the contrastive learning model are further used to evaluate the abnormality of each node with statistical estimation. In this way, the learning model is trained by a specific anomaly detection-aware target. Furthermore, since the input of the graph neural network module is batches of instance pairs instead of the full network, our framework can adapt to large networks flexibly. Experimental results show that our proposed framework outperforms the state-of-the-art baseline methods on all seven benchmark datasets.
\end{abstract}


\IEEEpeerreviewmaketitle

\section{Introduction}\label{sec:intro}

\IEEEPARstart{A}{ttributed} networks (a.k.a. attributed graphs), where nodes with attributes indicate real-world entities and links indicate the relationship between entities, are ubiquitous in various scenarios, including finance (trading networks) \cite{liu2018heterogeneous}, social media (social networks) \cite{tang2009relational, zhang2019your}, and e-commerce (item-user networks) \cite{ying2018graph, fan2019graph}. To utilize attributed network data to solve practical problems, a wide variety of graph analysis tasks have attracted significant research interests in recent years, such as node classification \cite{gcn_kipf2017semi, gat_ve2018graph}, graph classification \cite{gin_xu2019powerful, ying2018hierarchical}, and link prediction \cite{zhang2017weisfeiler, kipf2016variational}. Among these tasks, anomaly detection task on attributed networks is a vital research problem. Aiming to detect the instances that significantly deviate from the majority of instances \cite{pang2020deep} (in attributed networks, the data instances are nodes generally), anomaly detection has significant implications in many security-related applications, e.g., fraud detection and social spam detection \cite{dominant_ding2019deep}.

\begin{figure}[t]
	\centering
	\subfigure[Structural Anomaly]{
		\includegraphics[scale=0.36]{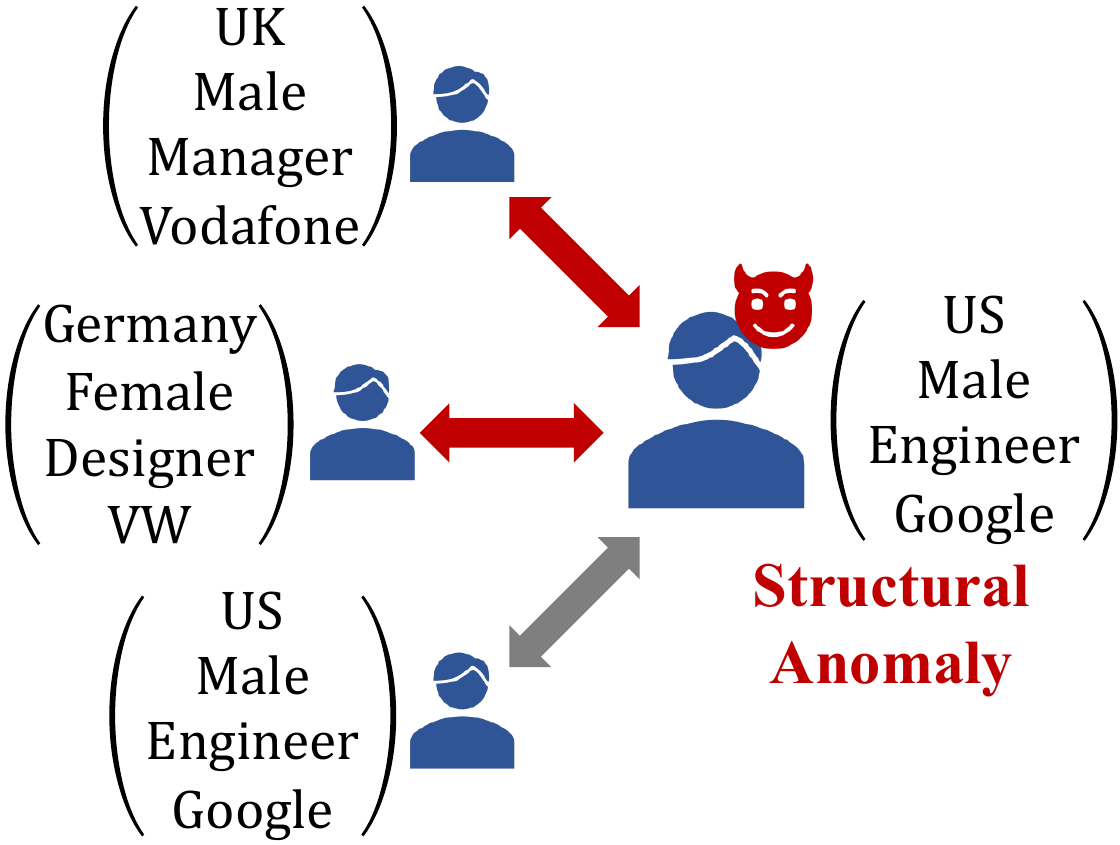}
	}
	\subfigure[Contextual Anomaly]{
		\includegraphics[scale=0.36]{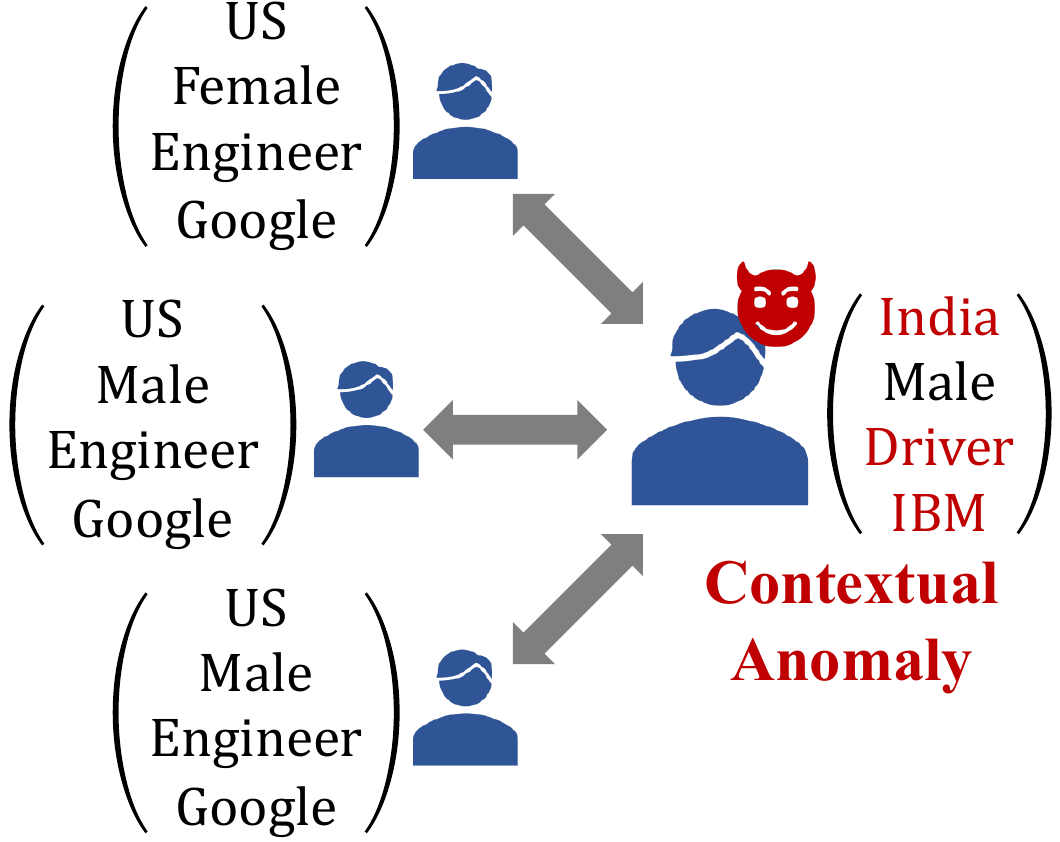}
	}
	
	\caption{Toy examples to illustrate different types of anomalies in attributed networks. A structural anomaly often has wrong links with other nodes but has normal attributes. For example, in Subfigure (a), an American engineer has a very low probability of being associated with a German designer as well as a British manager, so the links between them are abnormal. A contextual anomaly usually has a natural neighboring structure but its attributes are corrupted. For instance, in Subfigure (b), the attribute vector of the anomaly node is disturbed by noisy information, e.g., mismatched location, employer and occupation.} 
	\label{fig:anomaly}
\end{figure}

However, detecting anomalies effectively on attributed networks is not trivial due to the diversity of anomalies and the lack of supervision. Since attributed networks have both attribute information as well as structural information, they usually contain different types of anomalies. Figure \ref{fig:anomaly} provides an example to illustrate two basic types of anomalies: structural anomaly and contextual anomaly. The attribute information of the structural anomalies is often normal, while they have several abnormal links to other nodes. The contextual anomalies, differently, have natural neighboring structures but their attributes are corrupted (noisy or entirely different from all neighbors).
Such diversity makes it difficult to apply anomaly detection methods for attribute-only data (e.g., OC-SVM \cite{ocsvm_chen2001one}) or plain networks (e.g., LOF \cite{scan_xu2007scan}) to attributed networks directly. Therefore, an efficient anomaly detection approach should consider multiple patterns of anomalies. 
Moreover, resulting from the prohibitive cost for accessing ground-truth labels of anomalies, anomaly detection on attributed networks is predominately carried out in an unsupervised manner \cite{dominant_ding2019deep, amen_perozzi2016scalable}. That is to say, the algorithm has to conclude the normal pattern of data from the corrupted networks without supervision. Hence, a key is to fully and reasonably exploit existing information from attributed network data.

Recently, various methods have been proposed to deal with the anomaly detection task for attributed networks. The shallow methods, including AMEN \cite{amen_perozzi2016scalable}, Radar \cite{radar_li2017radar} and ANOMALOUS \cite{anomalous_peng2018anomalous}, leverage shallow learning mechanisms (e.g. ego-network analysis, residual analysis or CUR decomposition) to detect anomalies. Unfortunately, these models cannot fully address the computational challenge on attributed networks and fail to capture the complex interactions between different information modalities due to limitations of shallow mechanisms, especially when the feature is high-dimensional \cite{dominant_ding2019deep}. With the rocketing growth of deep learning for anomaly detection \cite{pang2020deep, pang2020deep2, pang2019deep, ruff2018deep}, researchers also present deep neural networks-based methods to solve the anomaly detection problem on attributed networks. DOMINANT \cite{dominant_ding2019deep} is one of the representative methods. It constructs a graph autoencoder to reconstruct the attribute and structure information simultaneously, and the abnormality is evaluated by reconstruction error. SpecAE \cite{specae_li2019specae} also leverages graph autoencoder to extract low-dimensional embedding, and carries out detection via density estimation. 

Although existing deep learning-based methods \cite{dominant_ding2019deep, specae_li2019specae} have achieved considerable performance for anomaly detection on graphs, they still have several shortcomings, largely attributed to the autoencoder backbone in their architectures. First, autoencoders aim to learn the latent representation by reconstructing the original data instead of detecting the anomaly itself. Although the anomaly scores can be computed according to reconstruction errors \cite{dominant_ding2019deep}, this kind of methods can only achieve suboptimal performance due to the fact that they do not target directly the anomaly detection objective. Second, autoencoder-based methods may not able to fully exploit the rich information of the attributed graph for effective graph representation learning.
Specifically, autoencoders simply rebuild the original data and they do not have any refinement for data. However, recent works \cite{cpc_oord2018representation, simclr_chen2020simple, dim_hjelm2018learning} have shown that more useful information can be mined in an unsupervised way if we design certain pretext tasks carefully based on augmented data. 
Third, graph autoencoder is the bottleneck to carry out anomaly detection on large-scale networks. Generally, the graph convolution operation in graph autoencoder needs to input and reconstruct the full networked data, which is unfeasible due to the explosive memory requirements when the network is large. 

As an alternative unsupervised learning technique, self-supervised contrastive learning is a promising solution to address the aforementioned limitations. By learning to contrast the elaborate instance pairs, the model can acquire informative knowledge without manual labels. Contrastive self-supervised learning has nice properties for anomaly detection task.
\textit{First}, contrastive learning mainly studies the matching of pairs of instances, which offers helpful information for anomaly detection. For the normal instance in graphs, there is a potential matching pattern between each node and its neighbors, e.g., the homophily hypothesis.
The anomalies, on the opposite, often present when there is an inconsistency/mismatch between attributes and structure, which violates the original matching pattern of networks. 
Moreover, different types of anomalies have different manners of mismatching: in Figure \ref{fig:anomaly}, the structural anomaly has individual abnormal links with uncorrelated nodes, which is partial inconsistency; the contextual anomaly, differently, has mismatched attributes with all neighbors. 
Contrastive learning, naturally, is capable to learn the matching patterns and capture various mismatching patterns via its intrinsic discriminative mechanism.
\textit{Second}, contrastive learning models provide a specific predicted score to measure the agreement between the elements in each instance pair, and the scale is highly related to the abnormality of instance. Since anomaly detection methods usually output a list of scores or a ranking to represent the abnormality of each node, the predicted scores of contrastive learning model can be utilized for anomaly detection directly. In this way, we can train the model via an objective that is highly relevant to anomaly detection.

In this paper, we propose a novel \underline{\textbf{Co}}ntrastive self-supervised \underline{\textbf{L}}earning framework for \underline{\textbf{A}}nomaly detection on attributed networks (\textbf{CoLA} for abbreviation). 
By sampling the well-designed instance pairs from the full network and using them to train the contrastive learning model, the information of network is exploited better. Concretely, our framework focuses on modeling the relationship between each node and its partial neighboring substructure, which can expose the various type of anomalies within networks.
Meanwhile, our CoLA framework is trained with a direct target to assist the anomaly detection task. We set the learning objective of our model to discriminate the agreement between the elements within the instance pairs, and the results can be further used to evaluate the abnormality of nodes. 
Besides, by splitting the network into separated lightweight instance pairs, our anomaly detection framework is compatible with large-scale networks. Specifically, our framework does not need to run graph convolution on full networks, so it successfully avoids the memory explosion problem.
To summarize, the main contributions are as follows:

\begin{itemize}
	\item We propose a contrastive self-supervised learning framework, CoLA, for the anomaly detection problem on attributed networks. To the best of our knowledge, this is the first contrastive self-supervised learning-based method for graph anomaly detection.
	\item We present a novel type of contrastive instance pair, ``target node v.s. local subgraph'', for attributed networks to adapt to the anomaly detection task, which efficiently captures the local information of a node and its neighboring substructure. 
	\item We design a contrastive learning model to learn the representative information from the node-subgraph instance pairs and provide discriminative scores for abnormality ranking. The proposed learning model is friendly to large-scale networked data.
	\item We conduct extensive experiments on various datasets to demonstrate the effectiveness of CoLA and its superiority compared with a range of baseline methods.
\end{itemize}

The rest of this paper is organized as follows. In Section \ref{sec:rw}, we first review the related works. Then, the preliminary definitions and notations are introduced in Section \ref{sec:definition}. Section \ref{sec:model} illustrates the overall pipeline and the components of our framework in detail. After that, we analyze the experimental results in Section \ref{sec:experiments} and then conclude our work in section \ref{sec:conclusion}.

\section{Related Work}\label{sec:rw}

In this section, we introduce the most related work: network embedding and graph neural networks, anomaly detection on attributed networks, and contrastive learning.

\subsection{Network Embedding and Graph Neural Networks}\label{subsec:rw_gnn}

Network embedding aims to embed nodes into latent vector spaces, where the inherent properties of the graph are preserved. For attributed network, the learned embedding should contain both structural and semantic information. For instance, SNE \cite{Liao2017Attributed} employs neural networks to model the interrelations between structure and attribute. TriDNR \cite{Pan2016Tri} jointly learns node embedding via tri-party information sources including node's structure, attributes and labels. NETTENTION \cite{yu2019self} leverages adversarial training mechanism and self-attention module to learn informative node embeddings.

Graph Neural Networks (GNNs) are a family of deep neural networks \cite{dl_lecun2015deep} for modeling the underlying relationships of non-Euclidean networks/graphs data \cite{gnn_survey_wu2020comprehensive}. 
The concept of GNN was firstly outlined in \cite{gori2005new}. After that, a series of spectral-based GNNs is proposed \cite{bruna2013spectral, henaff2015deep, defferrard2016convolutional}, which employs filters from the perspective of graph signal processing \cite{shuman2013emerging}. 
GCN \cite{gcn_kipf2017semi} performs a localized first-order approximation of spectral graph convolutions to learn node representation efficiently. 
GAT \cite{gat_ve2018graph} introduces the attention mechanism \cite{vaswani2017attention} to aggregate neighbors' information with adaptive weights. 
Some recent works try to improve GNNs in different directions, such as simplifying the computational complexity \cite{sgc_wu2019simplifying, nt2019revisiting}, training with adversarial scheme \cite{pan2019learning}, applying to large-scale graphs \cite{sage_hamilton2017inductive, clustergcn_chiang2019cluster, graphsaint_Zeng2020GraphSAINT}, and introducing novel operators \cite{gin_xu2019powerful, diffusion_klicpera2019diffusion}. 
Currently, GNNs have been applied to various research fields, such as time-series prediction \cite{wu2019graph, wu2020connecting}, hyperspectral image classification \cite{wan2020hyperspectral} and knowledge graph \cite{wan2020reasoning, xian2019reinforcement}.

In our proposed CoLA, GNN is a significant component of the contrastive learning model. We select GCN as the backbone of our GNN module. Flexibly, the GNN module in our framework can be set to any type of the aforementioned GNNs.

\subsection{Anomaly Detection on Attributed Networks}\label{subsec:rw_ad}

Anomaly detection on attributed networks attracts considerable research interest in recent years due to the wide application of attributed networks in modeling complex systems \cite{pang2020deep}. 
AMEN \cite{amen_perozzi2016scalable} detects anomalies by leveraging ego-network information of each node on attributed networks. Radar \cite{radar_li2017radar} characterizes the residuals of attribute information and its coherence with network information for anomaly detection. Further, ANOMALOUS \cite{anomalous_peng2018anomalous} jointly considers CUR decomposition and residual analysis for anomaly detection on attributed networks. Zhu et al. \cite{mixedad_zhu2020mixedad} present a joint learning model to detect mixed anomaly by core initiating and expanding. Despite their success on low-dimensional attributed network data, these methods cannot work well when the networks have complex structures and high-dimensional attributes due to the limitation of their shallow mechanisms.

With the rocketing growth of the deep learning technique \cite{dl_lecun2015deep}, several deep approaches are presented to solve the anomaly detection problem for attributed networks. DOMINANT \cite{dominant_ding2019deep} constructs an autoencoder with GCN layers to reconstruct both the attribute matrix and adjacency matrix. It defines the anomaly score of node as the weighted sum of its reconstruction errors of attribute and  structure. SpecAE \cite{specae_li2019specae} leverages a spectral graph autoencoder to extract the latent embedding of each node and uses Gaussian Mixture Model (GMM) to perform the detection. For dynamic networks, NetWalk \cite{yu2018netwalk} learns dynamically network representations with random walk sampling and autoencoder model, and detects anomalies with a clustering-based detector.

The above deep methods achieve superior performance over the shallow methods by introducing deep neural networks, but also have several shortcomings caused by their reconstruction mechanism with autoencoders. Firstly, reconstruction is a naive unsupervised learning solution that fails to make full use of data. On contrary, CoLA better utilizes the attribute and structure information in a self-supervised manner. Secondly, their reconstructive optimization target is not associated with anomaly detection. In contrast to them, our learning objective is to discriminate the agreement between nodes and subgraphs, which can indicate the abnormality of nodes directly. Thirdly, these methods require full adjacency and attribute matrix as model's input, which makes these algorithms unable to be run on large-sized network data due to the explosive memory requirements. In contrast, our framework learns with sampled instance pairs rather than the full network, which makes it flexibly adapt to large-scale networks.

\subsection{Contrastive Self-Supervised Learning}\label{subsec:rw_cl}

Contrastive self-supervised learning is a significant brunch of self-supervised learning \cite{cl_survey_liu2020selfsupervised}. Through handcrafted contrastive pretext tasks, these approaches learn representations by contrasting positive instance pairs against negative instance pairs \cite{simclr_chen2020simple}. 
Deep InfoMax \cite{dim_hjelm2018learning} learns the embedding of images by maximizing the mutual information (MI) between a local patch and its global context. 
As a follower, CPC \cite{cpc_oord2018representation} applies contrastive learning to speech recognition by maximizing the association between a segment of audio and its context audio. 
MoCo \cite{moco_he2020momentum} constructs a momentum encoder with a momentum-updated encoder to generate contrastive embedding. 
SimCLR \cite{simclr_chen2020simple} leverages different combinations of augmentation methods to build pair-wise samples. 
More recently, BYOL \cite{grill2020bootstrap} presents to contrast the representation of online network with target network, where the two networks are learned mutually. 
Simsiam \cite{chen2020exploring} adopts Siamese networks to learn visual representation learning in a contrastive manner.

Some recent works also exploit contrastive methods to graph learning. 
DGI \cite{dgi_velickovic2019deep} considers a node's representation and graph-level summary vector obtained by a readout function as a contrastive instance pair, and generates negative samples with graph corruption.
On the basis of DGI, Hassani et al. \cite{mtv_hassani2020contrastive} suggest a multi-view contrastive learning framework by viewing the original graph structure and graph diffusion \cite{diffusion_klicpera2019diffusion} as two different views. 
GCC \cite{gcc_qiu2020gcc} pre-trains GNN for universe graph data by sampling two subgraphs for each node as a positive instance pair and uses InfoNCE loss \cite{cpc_oord2018representation} to learn. 
GMI \cite{peng2020graph} considers maximizing the agreement between node's embedding and raw features of its neighbors as well as which between embedding of two adjacent nodes.

However, most of these works aim to learn data representation instead of detecting anomalies. To adapt to anomaly detection, our proposed CoLA framework has essential differences in both motivation and implementation compared with the aforementioned approaches. 
From the perspective of motivation, these approaches only take the embedding module of contrastive models as an encoder, and the discriminator module becomes useless when testing. Our proposed CoLA, in contrast, leverages the whole contrastive model to compute the anomaly scores for each node. 
From the perspective of implementation, since the existing instance pair definition cannot effectively capture the abnormality of nodes, we design a novel type of contrastive instance pair for graph contrastive learning, which pays close attention to the local information of each node rather than the global property. 

\section{Problem Formulation}\label{sec:definition}

\begin{table}[t]
	\small
	\centering
	\caption{Notations and explanations related to CoLA framework. The four blocks of the table (from top to bottom) show the notation of variables about attributed networks, contrastive learning, GNN, and CoLA's hyper-parameters respectively.} 
	\begin{tabular}{ p{66 pt}<{\centering} | p{155 pt}}  
		\toprule[1.0pt]
		Notation & Explanation  \\
		\cmidrule{1-2}
		$\mathcal{G}=(\mathcal{V}, \mathcal{E}, \mathbf{X})$ & A weighted attributed network \\
		$\mathcal{V}$ & The node set of $\mathcal{G}$ \\
		$\mathcal{E}$ & The edge set of $\mathcal{G}$ \\
		$\mathbf{X} \in \mathbb{R}^{n \times f}$ & The attribute matrix of $\mathcal{G}$ \\
		$\mathbf{A} \in \mathbb{R}^{n \times n}$ & The adjacency matrix of $\mathcal{G}$\\
		$\mathbf{x}_i \in \mathbb{R}^{f}$ & The attribute vector of the $i^{th}$ node in $\mathcal{G}$ \\
		$n$ & The number of nodes in $\mathcal{G}$ \\
		$f$ & The dimension of attributes in $\mathcal{G}$ \\
		$k_i$ & The anomaly score of the $i^{th}$ node in $\mathcal{G}$ \\
		\cmidrule{1-2}
		$P_i = (v_{i}, \mathcal{G}_i, y_i)$ & A contrastive instance pair with index $i$ \\
		$v_{i}$ & The target node of instance pair $P_i$ \\
		$\mathcal{G}_i$ & The local subgraph of instance pair $P_i$ \\
		$y_i \in \{0,1\}$ & The label of instance pair $P_i$ \\
		$\mathbf{A}_i \in \mathbb{R}^{c \times c}$ & The adjacency matrix of $\mathcal{G}_i$ \\
		\cmidrule{1-2}
		$\mathbf{H}^{(\ell)}_i  \in \mathbb{R}^{c \times d^{(\ell)}}$ & The hidden representation matrix of $\mathcal{G}_i$ outputted by the $\ell^{th}$ layer of GNN module\\
		$\mathbf{z}^{(\ell)}_i \in \mathbb{R}^{d^{(\ell)}}$ & The hidden representation vector of $v_i$ outputted by the $\ell^{th}$ layer of GNN module\\
		$\mathbf{E}_i \in \mathbb{R}^{c \times d}$ & The embedding matrix of the nodes in $\mathcal{G}_i$ \\
		$\mathbf{e}^{lg}_i \in \mathbb{R}^{d}$  & The embedding vector of $\mathcal{G}_i$\\
		$\mathbf{e}^{tn}_i \in \mathbb{R}^{d}$  & The embedding vector of $v_i$\\
		$s_i$  & The predicted score of $P_i$\\	
		$\mathbf{W}^{(\ell)} \in \mathbb{R}^{d^{(\ell-1)} \times d^{(\ell)}}$ & The learnable parameter of the $\ell^{th}$ layer of GNN module\\
		$\mathbf{W}^{(d)} \in \mathbb{R}^{d \times d}$& The learnable parameter of discriminator module\\
		\cmidrule{1-2}
		$R$  & The sampling rounds to calculate anomaly scores \\
		$c$  & The number of nodes within the local subgraphs \\
		$d$  & The dimension of embedding \\
		
		\bottomrule[1.0pt]
	\end{tabular}
	\label{table:notation}
\end{table}

In this paper, we use bold lowercase letters (e.g. $\mathbf{x}$), bold uppercase letters (e.g. $\mathbf{X}$), and calligraphic fonts (e.g. $\mathcal{V}$) to denote vectors,  matrices and sets, respectively. Accordingly, the definition of attributed networks is given as follows:

\begin{definition}
	\textbf{Attributed networks.} An attributed network can be denoted as $\mathcal{G}=(\mathcal{V}, \mathcal{E}, \mathbf{X})$, where $\mathcal{V}=\{v_1,\dots,v_n\}$ is the set of nodes ($\lvert \mathcal{V} \rvert = n$), $\mathcal{E}$ is the set of edges ($\lvert \mathcal{E} \rvert = m$), and $\mathbf{X} \in \mathbb{R}^{n \times f}$ is the attribute matrix. The $i^{th}$ row vector $\mathbf{x}_i \in \mathbb{R}^{f}$ of the attribute matrix denotes the attribute information of the $i^{th}$ node. An binary adjacency matrix $\mathbf{A} \in \mathbb{R}^{n \times n}$ is employed to denote the structure information of the attributed network, where $\mathbf{A}_{i,j} = 1$ if there is a link between nodes $v_i$ and $v_j$, otherwise $\mathbf{A}_{i,j} = 0$. Since the information of $\mathcal{V}$ and $\mathcal{E}$ is both contain by $\mathbf{A}$, an attributed network can also be denoted as $\mathcal{G}=(\mathbf{A}, \mathbf{X})$.
\end{definition}

With the aforementioned notations, the anomaly detection problem on attributed network can be formally stated as a ranking problem:

\begin{definition}
	\textbf{Anomaly Detection on Attributed networks.} Given an attributed network $\mathcal{G}=(\mathcal{V}, \mathcal{E}, \mathbf{X})$ with nodes $v_1,\dots,v_n$, the goal is to learn an anomaly score function $\mathit{f}$ to calculate the anomaly score $k_i = \mathit{f}(v_i)$ of each node. The anomaly score $k_i$ can represent the degree of abnormality of node $v_i$. By ranking all the nodes with their anomaly scores, the anomaly nodes can be detected according to their positions. 
\end{definition}

In this paper, we consider the setting of unsupervised anomaly detection, which is generally adopted by the previous works \cite{dominant_ding2019deep, radar_li2017radar, anomalous_peng2018anomalous}. In this setting, only the attributed network $\mathcal{G}$ that contains anomaly nodes is given, and neither the category label nor the abnormality label of the nodes is given in the training phase. 

For the convenience of the reader, the notations used in the paper are summarized in Table \ref{table:notation}.

\section{Methodology}\label{sec:model}

\begin{figure*}[htbp]
	\centering
	\includegraphics[width=1.0\textwidth]{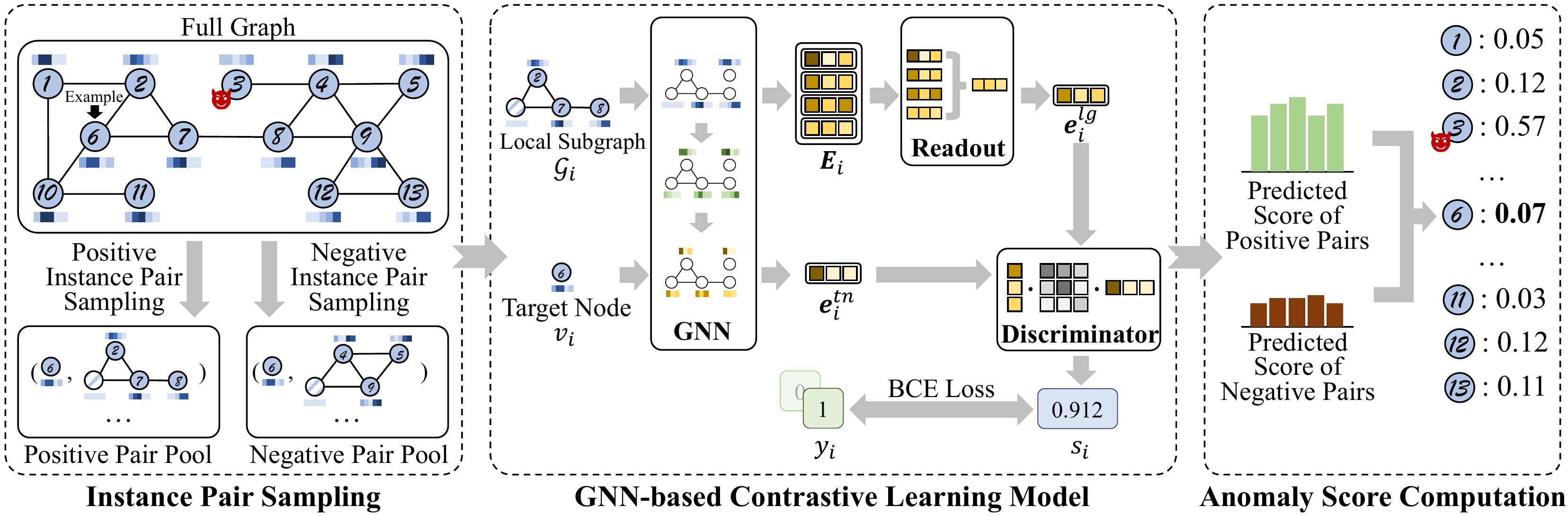}
	\caption{The overall framework of CoLA. The framework is composed of three components: instance pair sampling, GNN-based contrastive learning model, and anomaly score computation. Here we assume node $v_3$ is an anomaly since it has corrupted attributes, and the rest nodes are normal nodes. We take the abnormality estimation for node $v_6$ as an example. First of all, multiple positive and negative instance pairs are sampled, where $v_6$ is the target node. After that, the GNN-based contrastive learning model evaluates the predicted score for each instance pair. Finally, the anomaly score of $v_6$ is estimated by the statistical estimation of predicted scores of positive/negative pairs.}
	\label{fig:framework}
\end{figure*}

In this section, we describe the general framework of our proposed CoLA, as shown in Figure \ref{fig:framework}. CoLA on the highest level consists of three components, namely \textit{instance pair sampling}, \textit{GNN-based contrastive learning model}, and \textit{anomaly score computation}. 
At first, to generate the basic learning samples for contrastive self-supervised learning, we execute \textit{instance pair sampling} to sample the well-designed ``target node v.s. local subgraph'' instance pairs with a local substructure-based sampling strategy. Our designed instance pairs can take full advantage of the original data by paying close attention to each node and its local neighbors. 
After that, the \textit{GNN-based contrastive learning model} extracts the low-dimensional embeddings for target node and local subgraph with the GNN and readout module and then calculates a discriminative score for each instance pair with a discriminator. The predicted score that evaluates the agreement between the target node and subgraph can further indicate the abnormality of the corresponding target node. As such, the training of the model is guided by an objective that relates to the anomaly detection task. 
The final step, \textit{anomaly score computation}, is to measure the abnormalities of all nodes with anomaly scores, so we can pick the anomalies out by ranking the scores. Specifically, the anomaly scores for each node are calculated by the predicted scores of positive/negative pairs which are acquired by multi-round sampling. The detection results can be regarded as an expectation of multiple times of observation for the compatibility between each node and its local substructure. 
In the rest of this section, we introduce the three main components of our framework in detail (Subsection \ref{subsec:model_instances} to \ref{subsec:model_anoscore}). Then we describe the overall pipeline and algorithm of CoLA in Subsection \ref{subsec:model_framework}. In Subsection \ref{subsec:model_complex}, the time complexity of our framework is analyzed.

\subsection{Contrastive Instance Pair Definition}\label{subsec:model_instances}

The success of contrastive learning frameworks largely relies on the definition of the contrastive instance pair. Unlike computer vision or natural language processing tasks where an instance can be defined as an image or a sentence straightforwardly, the instance definition in graphs is not so clear as well \cite{gcc_qiu2020gcc}. Some previous works have defined different types of instance pair in graphs, such as ``full graph v.s. node'' \cite{dgi_velickovic2019deep}, ``large subgraph v.s. node'' \cite{mtv_hassani2020contrastive} and ``subgraph v.s. subgraph'' \cite{gcc_qiu2020gcc}. However, none of them is designed or optimized for the anomaly detection task. Since the abnormality of a node is usually related to the relationship between the node itself and its neighboring structure, we should design a novel type of contrastive instance pair to capture such local property.

To model the local distribution pattern of nodes in a network, our definition of contrastive instance pair focuses on the relationship between a node and its enclosing substructure. Specifically, we design a  ``target node v.s. local subgraph'' instance pair for anomaly detection on attributed networks. The first element of the instance pair is a single node, which is named ``target node'' in our framework. The target node can be set as any node in the network. The second element of the instance pair is a local subgraph that sampled from an initial node. For positive instance pairs, the initial node is set as the target node, then the sampled subgraph is composed of the neighbor nodes of the target node. For negative instance pairs, the initial node is randomly selected from all nodes except the target node. As a result, there is mismatching between the target node and the local subgraph in negative pairs. 

The main motivation for such design is that the anomalies in attributed networks are usually reflected in the inconsistency between node and its local neighbors, while the global information is often independent of anomalies. As is shown in Figure \ref{fig:anomaly}, both types of anomaly nodes have a mismatch with their near neighbors. Our design purposefully focuses on picking out such mismatch by learning the ``node-local subgraph'' matching pattern. Differently, the existing works (e.g. DGI \cite{dgi_velickovic2019deep}) mainly consider the global property of nodes, which is helpful for network embedding but has a minor contribution on detecting anomalies. The comparison experiments in Section \ref{subsec:exp_ad_result} illustrates that our designed instance pair is critical to capture anomaly.

\begin{figure}[htbp]
	\centering
	\includegraphics[width=0.4\textwidth]{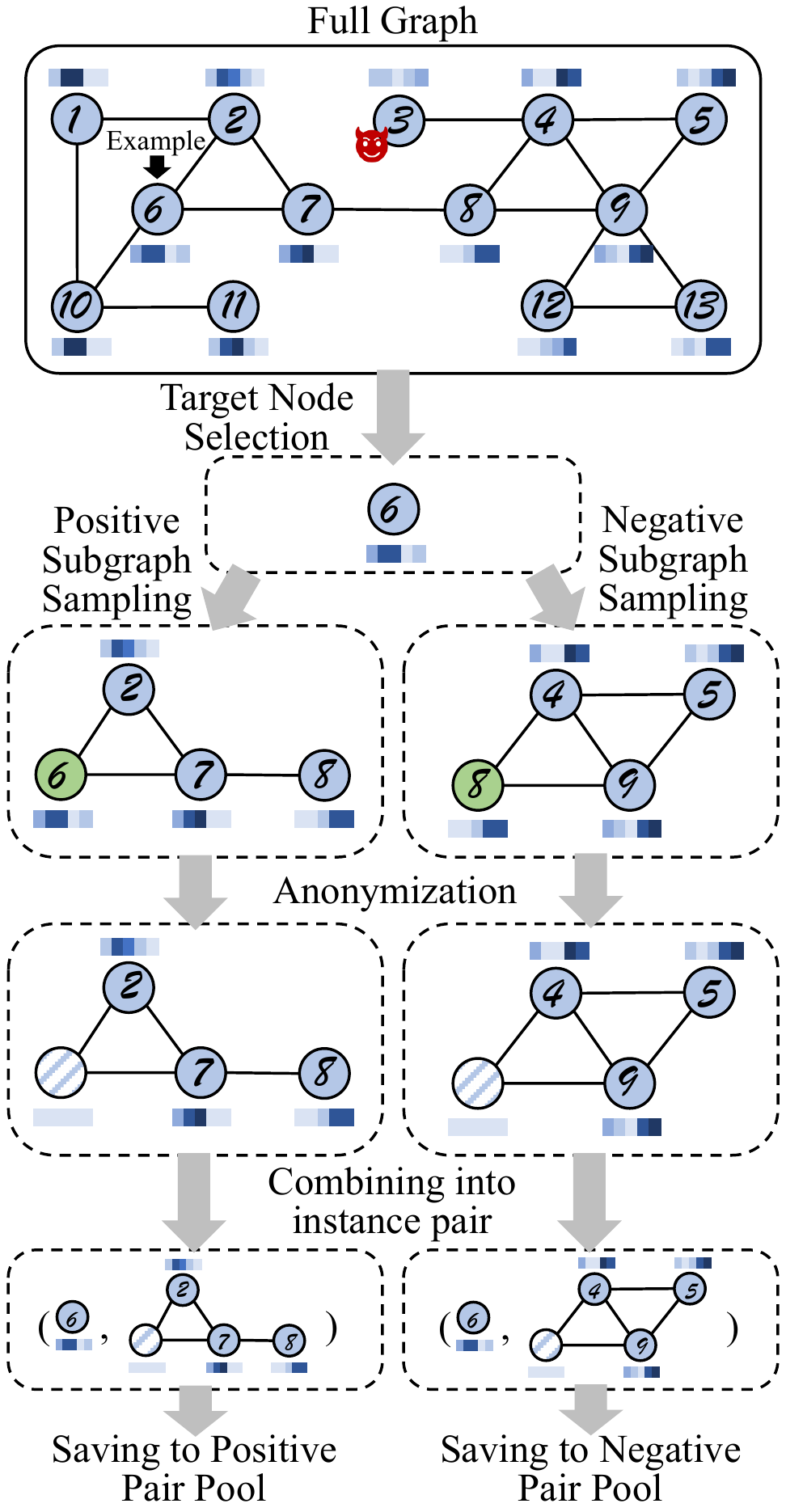}
	\caption{The sampling process of contrastive instance pairs. Here we select node $v_6$ as the example of target node. The initial nodes for subgraph sampling are marked in green. The blue-white stripe means the embedding of the corresponding node is masked with zero vector.}
	\label{fig:sampling}
\end{figure}

As shown in Figure \ref{fig:sampling}, a simple sketch map is used to demonstrate the sampling process of our proposed contrastive instance pairs. The sampling follows four steps: target node selection, subgraph sampling, anonymization, and combining into instance pair. 

\begin{enumerate}[(1)]
	\item  \textbf{Target node selection.} A single target node needs to be specified first. In practice, we traverse each node in the graph as the target node in random order within an epoch. Therefore, the target node selection can be viewed as a stochastic sampling without replacement.
	
	\item  \textbf{Subgraph sampling.} A local subgraph is defined as the adjacent substructure near an initial node. As we mentioned above, we sample the subgraph for positive pair and negative pair by setting the initial node as the target node and a randomly sampled node respectively. Inspired by \cite{gcc_qiu2020gcc}, we adopt random walk with restart (RWR) \cite{rwr_tong2006fast} as local subgraph sampling strategy due to its usability and efficiency. Other graph sampling algorithms such as forest fire \cite{ff_leskovec2006sampling} are also available in our framework. 
	
	\item  \textbf{Anonymization.} The purpose of anonymization is to prevent the contrastive learning model from easily identifying the existence of target nodes in local subgraphs. Concretely, we set the attribute vectors of the initial nodes into zero vectors. As such, the information of target nodes is hidden.
	
	\item  \textbf{Combining into instance pair.} The final step is to combine the target node and relevant subgraphs into instance pairs. After combination, the positive pair and negative pair are saved to corresponding sample pools respectively.
\end{enumerate}

\subsection{GNN-based Contrastive Learning Model}\label{subsec:model_model}

The sampled instance pairs are used to train the GNN-based contrastive learning model. For instance pair $P_i$ with its label, the containing data can be denoted as:

\begin{equation}
\label{eq:pair_definition}
P_i = (v_{i}, \mathcal{G}_i, y_i),
\end{equation}

\noindent where $v_i$ is the target node whose attribute vector is $\mathbf{x}_{v_i}$, $\mathcal{G}_i$ is the local subgraph which can be denoted as: 

\begin{equation}
\label{eq:local_subgraph}
\mathcal{G}_i = (\mathbf{A}_i, \mathbf{X}_i),
\end{equation}

\noindent and $y_i$ is the label of $P_i$ which can be denoted as:

\begin{equation}
\label{eq:label_definition}
y_i=\left\{
\begin{aligned}
1,  & \text{ $P_i$ is a positive instance pair} \\
0,  & \text{ $P_i$ is a negative instance pair}
\end{aligned}
.
\right.
\end{equation}

As is demonstrated in the middle part of Figure \ref{fig:framework}, our proposed GNN-based contrastive learning model is composed of three main components: GNN module, readout module, and discriminator module. 

\subsubsection{GNN module}
The target of the GNN module is to aggregate the information between nodes in the local subgraph and transfer the high-dimensional attributes into a low-dimensional embedding space. The local subgraph $\mathcal{G}_i$ is fed into a GNN with multiple stacked layers, where a single layer can be written as:

\begin{equation}
\label{eq:gnn}
\mathbf{H}^{(\ell)}_i = GNN(\mathbf{A}_i,\mathbf{H}^{(\ell-1)}_i; W^{(l-1)}),
\end{equation}

\noindent where $\mathbf{H}^{(\ell-1)}_i$ an $\mathbf{H}^{(\ell)}_i$ is the hidden representation matrices learned by the $(\ell-1)$-th layer and $\ell$-th layer respectively, and  $W^{(\ell-1)}$ is the learnable parameter set of the $(\ell-1)$-th layer. With a $L$-layer GNN, the input representation $\mathbf{H}^{(0)}_i$ is defined as the attribute matrix $\mathbf{X}_i$, and the output representation $\mathbf{H}^{(L)}_i$ is the embeddings of subgraph nodes which is denoted as $\mathbf{E}_i$. $GNN(\cdot)$ can be set to any type of mainstream GNNs, such as GCN \cite{gcn_kipf2017semi}, GAT \cite{gat_ve2018graph} or GIN \cite{gin_xu2019powerful}. In practice, we adopt GCN due to its high efficiency. Then Equation (\ref{eq:gnn}) can be specifically written as:

\begin{equation}
\label{eq:gcn}
\mathbf{H}^{(\ell)}_i=\phi\left(\widetilde{\mathbf{D}}_i^{-\frac{1}{2}} \widetilde{\mathbf{A}}_i \widetilde{\mathbf{D}}_i^{-\frac{1}{2}} \mathbf{H}^{(\ell-1)}_i \mathbf{W}^{(\ell-1)}\right),
\end{equation}

\noindent where $\widetilde{\mathbf{A}}_i = \mathbf{A}_i + \mathbf{I}$ is the subgraph adjacency matrix with self-loop, $\widetilde{\mathbf{D}}_i$ is the degree matrix of local subgraph, $\mathbf{W}^{(\ell-1)} \in \mathbb{R}^{d^{(\ell-1)} \times d^{(\ell)}}$ is the weight matrix of the $(\ell-1)$-th layer, $\phi(\cdot)$ is the activation function such as ReLU. 

In order to contrast them in the same feature space, besides the nodes in local subgraph, the target node should also be mapped to the same embedding space. Since there is no structure information with a single node, we only employ the weight matrices of GCN and corresponding activation function to transform the attributes of the target node. Concretely, such transformation can be viewed as a DNN: 

\begin{equation}
\label{eq:dnn}
\mathbf{z}^{(\ell)}_i=\phi\left(\mathbf{z}^{(\ell-1)}_i \mathbf{W}^{(\ell-1)}\right),
\end{equation}

\noindent where $\mathbf{z}^{(\ell-1)}_i$ and $\mathbf{z}^{(\ell)}_i$ are the hidden representation row vectors for target node that learned by the $(\ell-1)$-th layer and $\ell$-th layer respectively, and $\mathbf{W}^{(\ell-1)}$ is the weight matrix shared with GCN. The input $\mathbf{z}^{(0)}_i$ is defined as the attribute row vector of target node $\mathbf{x}_{v_i}$, and the output is marked as target node embedding $\mathbf{e}^{tn}_{i}$.

\subsubsection{Readout module}
The target of our readout module is to transfer the embeddings of nodes in subgraph $\mathbf{E}_i$ into a local subgraph embedding vector $\mathbf{e}^{lg}_i$. For simplification, we use the average pooling function as our readout function, which has been widely used in previous works \cite{mtv_hassani2020contrastive}. Specifically, the readout function is written as follows:

\begin{equation}
\label{eq:readout}
\mathbf{e}^{lg}_i = Readout(\mathbf{E}_i)  = \sum_{k=1}^{n_i} \frac{\left(\mathbf{E}_{i}\right)_{k}}{n_i},
\end{equation}

\noindent where $\left(\mathbf{E}_{i}\right)_{k}$ is the $k$-th row of $\mathbf{E}_i$, and $n_i$ is the number of nodes of the local subgraph $\mathcal{G}_i$.

\subsubsection{Discriminator module}
The discriminator module is the core component of our contrastive learning model. It contrasts the embeddings of the two elements in an instance pair and outputs the final predicted score. Here, we apply a simple bilinear scoring function, which is also employed by \cite{cpc_oord2018representation}. The predicted score can be calculated by:

\begin{equation}
\label{eq:discriminator}
s_i = Discriminator(\mathbf{e}^{lg}_i, \mathbf{e}^{tn}_i) = \sigma\left({\mathbf{e}^{lg}_i}  \mathbf{W}^{(d)} {\mathbf{e}^{tn}_i}^\top \right),
\end{equation}

\noindent where $\mathbf{W}^{(d)}$ is the weight matrix of discriminator, and $\sigma(\cdot)$ is the logistic sigmoid function.

\subsubsection{Objective function}
By integrating the aforementioned three components, our proposed GNN-based contrastive learning model can be considered as a binary classification model to predict the labels of contrastive instance pairs:

\begin{equation}
\label{eq:clm}
s_i = CLM(v_{i}, \mathcal{G}_i),
\end{equation}

\noindent where $CLM(\cdot)$ is the contrastive learning model.

Here, our objective is to make the predicted $s_i$ and the ground-truth label $y_i$ as close as possible. Therefore, we adopt a standard binary cross-entropy (BCE) loss, which is a common choice for binary classification problems, as our objective function. Its utility has been validated by other contrastive self-supervised learning works \cite{dim_hjelm2018learning, dgi_velickovic2019deep, mtv_hassani2020contrastive}. Concretely, for a batch of $P_i = (v_{i}, \mathcal{G}_i, y_i)$ with batch size $N$, the objective function is given as follows:

\begin{equation}
\label{eq:obj_func}
\begin{aligned}
\mathcal{L} &= -\sum_{i=1}^{N} y_{i} \log \left(s_{i}\right)+\left(1-y_{i}\right) \log \left(1-s_{i}\right) \\
&= -\sum_{i=1}^{N} y_{i}  \log \left(CLM(v_{i}, \mathcal{G}_i)\right)+ \\
& \quad \left(1-y_{i}\right) \log \left(1-CLM(v_{i}, \mathcal{G}_i)\right) .
\end{aligned}
\end{equation}

It should be noted that different from the softplus version BCE in \cite{dim_hjelm2018learning} and the label-balanced version BCE in \cite{dgi_velickovic2019deep, mtv_hassani2020contrastive}, a vanilla BCE is utilized here. The reason is that we execute a balanced sampling when we sample the positive and negative instance pairs. To adapt to more complex sampling strategies in further research, corresponding objective functions are also alternative in our framework.

\subsection{Anomaly Score Computation}\label{subsec:model_anoscore}

After the contrastive learning model is well trained, we acquire a classifier to discriminate the agreement between substructures and nodes. An ideal GNN model with an appropriate number of parameters would tend to learn the matching pattern of normal samples since they occupy the vast majority of the training data. For the anomalies, it is much harder to fit their pattern due to its irregularity and diversity. Under ideal conditions, for a normal node, the predicted score of its positive pair $s^{(+)}$ should be close to $1$, while the negative one $s^{(-)}$ should be close to $0$. For an anomalous node, the predicted scores of its positive and negative pairs would be less discriminative (close to $0.5$) because the model cannot well distinguish its matching pattern. Based on the above property, for each node $v_i$, we can simply define the anomaly score as the difference value between its negative and positive score:

\begin{equation}
\label{eq:simple_ano_func}
f(v_i)= s_{i}^{(-)}-s_{i}^{(+)}.
\end{equation}

However, a sampled local subgraph can only be viewed as a partial observation of the target node's neighboring structure, which cannot represent the whole neighbor distribution of the target node. Incomplete observation will lead to an incomplete perception of abnormality, which will affect the performance of anomaly detection. For example, some structural anomaly nodes have several abnormal links to uncorrelated nodes, while most of their neighbors are normal. Then, if we only estimate the abnormality with one-shot sampling, once the normal neighbors are sampled as the local subgraph, such abnormality will be ignored.

To solve this problem, we propose to use the predicted scores of multi-round and positive-negative sampling to generate anomaly scores. Specifically, for each node $v_i$ in the attributed network, we sample $R$ positive instance pairs as well as $R$ negative instance pairs via the sampling strategy introduced in Subsection \ref{subsec:model_instances}, and $R$ is the number of sampling round. These instance pairs are denoted as $(P_{i,1}^{(+)},\cdots, P_{i,R}^{(+)})$ and $(P_{i,1}^{(-)},\cdots, P_{i,R}^{(-)})$ respectively. Then, these instance pairs are fed into the contrastive learning model $CLM(\cdot)$ to calculate the predicted scores $(s_{i,1}^{(+)},\cdots, s_{i,R}^{(+)},s_{i,1}^{(-)}, \cdots, s_{i,R}^{(-)})$, which is computed via Equation (\ref{eq:clm}). Finally, the anomaly score of $v_i$ is obtained by computing the average value of multi-round differences between the scores of negative and positive pairs:

\begin{equation}
\label{eq:ano_func}
f(v_i)= \frac{\sum_{r=1}^{R}(s_{i,r}^{(-)}-s_{i,r}^{(+)})}{R},
\end{equation}

\noindent where $f(\cdot)$ is the anomaly score mapping function which is the ultimate goal of our anomaly detection framework. 

From a statistical perspective, executing the $R$ rounds sampling is to estimate the difference of normality between a node's neighboring substructure and remote substructure. In principle, the larger $R$ is, the more accurate the estimation is. In practice, we set $R$ as a hyper-parameter of our framework, and we discuss the selection of $R$ in Section \ref{subsec:exp_param_study}. 

\begin{algorithm}[t]
	\caption{The Overall Procedure of CoLA}
	\label{algo:cola}
	\renewcommand{\algorithmicrequire}{\textbf{Input:}}
	\renewcommand{\algorithmicensure}{\textbf{Output:}}
	\begin{algorithmic}[1]
		\REQUIRE {Attributed network: $\mathcal{G}=(\mathcal{V}, \mathcal{E}, \mathbf{X})$, Number of training epochs: $T$, Batch size: $B$, Number of sampling rounds: $R$}.
		\ENSURE {Anomaly score mapping function: $f(\cdot)$}.
		\STATE Randomly initialize the parameters of contrastive learning model $(\mathbf{W}^{(0)}, \cdots, \mathbf{W}^{(L)}, \mathbf{W}^{(d)})$ 
		\STATE $//$ {\it Training phase.}
		\FOR{$t \in 1,2,\cdots,T$}
		\STATE $\mathcal{B} \leftarrow$ (Randomly split $\mathcal{V}$ into batches of size $B$)
		\FOR{batch $b=(v'_{1},\cdots,v'_{B}) \in \mathcal{B}$}
		\STATE Sample positive instance pairs $(P_{1}^{(+)},\cdots, P_{B}^{(+)})$ where $(v'_{1},\cdots,v'_{B})$ are the target node respectively.
		\STATE Sample negative instance pairs $(P_{1}^{(-)},\cdots, P_{B}^{(-)})$ where $(v'_{1},\cdots,v'_{B})$ are the target node respectively.
		\STATE Calculate the predicted scores $(s_{1}^{(+)},\cdots, s_{B}^{(+)},s_{1}^{(-)},\cdots,$ $s_{B}^{(-)})$ of instance pairs $(P_{1}^{(+)},\cdots, P_{B}^{(+)},P_{1}^{(-)},\cdots,P_{B}^{(-)})$ via Equation (\ref{eq:clm}).
		\STATE Calculate $\mathcal{L}$ via Equation (\ref{eq:obj_func}).
		\STATE Back propagation and update the parameters of contrastive learning model $(\mathbf{W}^{(0)}, \cdots, \mathbf{W}^{(L)}, \mathbf{W}^{(d)})$.
		\ENDFOR
		\ENDFOR
		
		\STATE $//$ {\it Inference phase.}
		\FOR{$v_i \in \mathcal{V}$}
		\STATE Sample $R$ positive instance pairs $(P_{i,1}^{(+)},\cdots, P_{i,R}^{(+)})$ where $v_i$ is the target node.
		\STATE Sample $R$ negative instance pairs $(P_{i,1}^{(-)},\cdots, P_{i,R}^{(-)})$ where $v_i$ is the target node.
		\STATE Calculate the predicted scores $(s_{i,1}^{(+)},\cdots, s_{i,R}^{(+)},s_{i,1}^{(-)}, \cdots, s_{i,R}^{(-)})$ of instance pairs $(P_{i,1}^{(+)},\cdots, P_{i,R}^{(+)},P_{i,1}^{(-)}, \cdots, P_{i,R}^{(-)})$ via Equation (\ref{eq:clm})
		\STATE Calculate the anomaly score $f(v_i)$ via Equation (\ref{eq:ano_func}).
		\ENDFOR
	\end{algorithmic}
\end{algorithm}

Furthermore, computing the mean value is the simplest way to process multi-round results. Theoretically, there are more statistical properties that can be mined, such as variance, minimum/maximum value, and distribution property. However, in practice, we found that calculating the average value is the most effective solution compared with introducing the above factors. In spite of this situation, we still think that the further mining of statistical properties of the multi-round predicted scores is one of the potential directions in the future since different types of anomalies will show different characteristics of the score distribution.

\subsection{CoLA: An Anomaly Detection Framework} \label{subsec:model_framework}  

In this subsection, we introduce the overall pipeline of our proposed CoLA framework. The pipeline is divided into two phases: training phase and inference phase. In the training phase, the contrastive learning model is trained with sampled instance pairs in an unsupervised fashion. After that, the anomaly score for each node is obtained in the inference phase.

The overall procedure of our CoLA framework is depicted in Algorithm \ref{algo:cola}. In an epoch of the training phase, we first split the set of nodes $\mathcal{V}$ into several mini-batches. Then, in each iteration, a positive pair and a negative pair are sampled for each node in the current mini-batch. After that, the corresponding predicted scores are calculated with the instance pairs, then the BCE loss is computed. To optimize the parameters of contrastive learning model, a backpropagation is executed with a gradient descent algorithm. After the training phase, a multi-round positive-negative sampling procedure is carried out to generate anomaly scores. As described in Subsection \ref{subsec:model_anoscore},  for each node $v_i$, $R$ positive pairs and $R$ negative pairs are sampled, then they are fed into the well-trained model to calculate the predicted scores. Finally, the anomaly score is obtained via Equation (\ref{eq:ano_func}).

\textbf{Discussion on anomaly detection for large-scale networks.} As we introduced above, the contrastive learning model is trained by a mini-batch of instance pairs independently in an iteration. Meanwhile, the computation of anomaly scores is also completely independent. That is to say, the space complexity of CoLA is uncorrelated with the number of nodes $n$ at all. Such a nice property makes it possible to apply our proposed CoLA framework to large-scale networks. When the size of network is large ($n$ is large), we do not need to feed the full network into the GCN model, which is unfeasible due to the explosive need for space complexity \cite{clustergcn_chiang2019cluster, graphsaint_Zeng2020GraphSAINT}. Instead, in our framework, the full network is decomposed into instance pairs and all we need is to adjust the batch size and subgraph size to meet the memory constraint. 

\subsection{Complexity Analysis}\label{subsec:model_complex}

We analyze the time complexity of the proposed framework by considering the three main components respectively. For instance pair sampling, the time complexity of each RWR subgraph sampling is $\mathcal{O}(c\delta)$ ($\delta$ is the mean degree of network). In the inference phase, we run $R$ rounds of sampling for each node, then the total time complexity becomes $\mathcal{O}(cn\delta R)$. For the GNN-based contrastive learning model, the time complexity is mainly generated by the GNN module, which is $\mathcal{O}(c^2)$ for each pair and $\mathcal{O}(c^2nR)$ for a total. For anomaly score computation, the time complexity is far less than the above two phases, so here we ignore this term. To sum up, the overall time complexity of CoLA is $\mathcal{O}(cnR(c+\delta))$.

\section{Experiments}\label{sec:experiments}

In this section, we conduct experiments to show the effectiveness of CoLA framework. We first introduce the datasets used for experiments and experiments setup. Then, we demonstrate the experimental results including the comparison of performance, parameter study, and ablation study.

\subsection{Datasets} \label{subsec:exp_dataset}

We evaluate the proposed framework on seven widely used benchmark datasets for anomaly detection on attributed networks. The datasets include two social network datasets (BlogCatalog and Flickr) and five citation network datasets (Cora, Citeseer, Pubmed, ACM and ogbn-arxiv) \cite{sen2008collective, tang2009relational, tang2008arnetminer}. The statistics of these datasets are demonstrated in Table \ref{table:dataset}, and the detailed descriptions are given as follows:

\begin{itemize} 
	
	\item \textbf{Social Networks}. BlogCatalog and Flickr\footnote{http://socialcomputing.asu.edu/pages/datasets} \cite{tang2009relational} are two typical social network datasets acquired from the blog sharing website BlogCatalog and the image hosting and sharing website Flickr, respectively. In these datasets, nodes denote the users of websites, and links represent the following relationships between users. In social networks, users usually generate personalized content such as posting blogs or sharing photos with tag descriptions, thus these text contents are regarded as node attributes.

	\item \textbf{Citation Networks}. Cora, Citeseer, Pubmed\footnote{http://linqs.cs.umd.edu/projects/projects/lbc} \cite{sen2008collective}, ACM \cite{tang2008arnetminer}, and ogbn-arxiv\footnote{https://github.com/snap-stanford/ogb} \cite{ogb_hu2020ogb} are five available public datasets, which are composed of scientific publications. In these networks, nodes denote the published papers while edges represent the citation relationships between papers. For the first four datasets, the attribute vector of each node is the bag-of-word representation whose dimension is determined by the dictionary size. For ogbn-arxiv dataset, each node has a 128-dimensional attribute vector obtained by averaging the embeddings of words in the paper's title and abstract. Note that the ogbn-arxiv dataset is a \textbf{large-scale} graph dataset from Open Graph Benchmark (OGB) where over 169k nodes and 1.1m edges are contained.

\end{itemize}

\begin{table}
	\centering
	\caption{The statistics of the datasets. The upper two datasets are social networks, and the remainders are citation networks.}
	\begin{tabular}{@{}c|c|c|c|c@{}}
		\toprule
		\textbf{Dataset} &  \textbf{$\sharp$ nodes} & \textbf{$\sharp$ edges} & \textbf{$\sharp$ attributes} & \textbf{$\sharp$ anomalies} \\ \midrule
		\textbf{BlogCatalog}   & 5,196    & 171,743      & 8,189            & 300               \\
		\textbf{Flickr}             & 7,575    & 239,738     & 12,407          & 450               \\
		\textbf{ACM}              & 16,484  & 71,980       & 8,337            & 600               \\ 
		\textbf{Cora}              & 2,708    & 5,429        & 1,433            & 150                \\
		\textbf{Citeseer}         & 3,327    & 4,732        & 3,703            & 150                \\
		\textbf{Pubmed}         & 19,717   & 44,338      & 500               & 600                \\
		\textbf{ogbn-arxiv}         & 169,343   & 1,166,243      & 128               & 6000                \\		
		
		\bottomrule
	\end{tabular}
	\label{table:dataset}
\end{table}

Since there are no ground-truth anomalies in the aforementioned datasets, synthetic anomalies are needed to inject into the clean attributed networks for our evaluation. Following the previous researches \cite{song2007conditional, ding2019interactive, dominant_ding2019deep}, we inject a combined set of anomalies (i.e., structural anomalies and contextual anomalies) for each dataset:

\begin{itemize} 
	\item \textbf{Structural anomalies injection}. The structural anomalies are acquired by perturbing the topological structure of networks \cite{ding2019interactive}. Concretely, some small cliques composed of originally unrelated nodes are generated as anomalies. The intuition is that in a small clique, a small set of nodes are much more closely linked to each other than average, which can be regarded as a typical structural anomalous situation in real-world networks \cite{skillicorn2007detecting}. To make the cliques, we first specify the clique size $p$ and the number of cliques $q$. When generating a clique, we randomly choose $p$ nodes from the set of nodes $\mathcal{V}$ and make them fully connected. As such, the selected $p$ nodes are all marked as structural anomaly nodes. To generate $q$ cliques, we repeat the above process for $q$ times. Finally, a total of $p \times q$ structural anomalies were injected. According to the size of datasets, we control the number of injected anomalies. We fix $p=15$ and set $q$ to $10, 15, 20, 5, 5, 20, 200$ for BlogCatalog, Flickr, ACM, Cora, Citeseer, Pubmed and ogbn-arxiv, respectively.
	
	\item \textbf{Contextual anomalies injection}. Here, we create the contextual anomalies by perturbing the attribute of nodes following the schema introduced by \cite{song2007conditional}. When generate a single contextual anomaly node, we first randomly pick a node $v_i$ as the target, and then sample another $k$ nodes $\mathcal{V}^{(c)}=(v^{(c)}_{1}, \cdots, v^{(c)}_k)$ as a candidate set. After that, for each $v^{(c)} \in \mathcal{V}^{(c)}$, we calculate the Euclidean distance between its attribute vector $\mathbf{x}^{(c)}$ and $v_i$'s attribute vector $\mathbf{x}_i$. Then, we pick the node $v^{(c)}_j \in \mathcal{V}^{(c)}$ which has the largest Euclidean distance to $v_i$, and change $\mathbf{x}_i$ to $\mathbf{x}^{(c)}_i$. To balance the equal numbers of the two types of anomalies, we set the number of context anomalies as $p \times q$, which means the above operation is repeated for $p \times q$ times to generate all the contextual anomalies. Here, we set $k=50$ to ensure the disturbance amplitude is large enough.
	
\end{itemize}

Following the aforementioned injection methods, we finally obtain the perturbed networks, and the total number of anomalies is given in the last column of Table \ref{table:dataset}. All the category labels are removed in our experiments, and the anomalous labels are only visible in the inference phase.

\subsection{Experimental Settings} \label{subsec:exp_setting}  

In this subsection, we introduce the settings of our experiments, including baselines for comparison, metrics for evaluation, and parameter setting of our framework.

\subsubsection{Baselines}
We compare our proposed framework CoLA with five popular methods for anomaly detection or graph contrastive learning:

\begin{itemize} 
	\item \textbf{AMEN\footnote{https://github.com/phanein/amen} \cite{amen_perozzi2016scalable}}. AMEN is an ego-network analysis-based anomaly detection method. It identifies anomalies by evaluating the attribute correlation of nodes per ego-network.
	
	\item \textbf{Radar\footnote{http://people.virginia.edu/\%7Ejl6qk/code/Radar.zip} \cite{radar_li2017radar}}. Radar is a residual analysis-based method. It characterizes the residuals of attribute information and its coherence with network information for anomaly detection on networks.
	
	\item \textbf{ANOMALOUS\footnote{http://people.virginia.edu/\%7Ejl6qk/code/ANOMALOUS.zip} \cite{anomalous_peng2018anomalous}}. ANOMALOUS is also based on the residual analysis. It is a joint learning framework that conducts attribute selection and anomaly detection as a whole based on CUR decomposition and residual analysis.
	
	\item \textbf{DOMINANT\footnote{https://github.com/kaize0409/GCN\_AnomalyDetection} \cite{dominant_ding2019deep}}. DOMINANT is the state-of-the-art unsupervised anomaly detection framework based on deep learning. It utilizes a graph convolution autoencoder to jointly reconstruct the adjacency matrix as well as the attribute matrix. It evaluates the abnormality of each node by computing the weighted sum of the reconstruction error terms.
	
	\item \textbf{DGI\footnote{https://github.com/PetarV-/DGI} \cite{dgi_velickovic2019deep}}. DGI is a representative method for contrastive graph representation learning. It generates node embedding via node-graph contrasting. A bilinear function serves as a discriminator to predict the agreement between node and original/corrupted graphs.

\end{itemize}

\begin{figure*}[ht]
	\centering
	\subfigure[BlogCatalog]{
		\includegraphics[width=4.5cm]{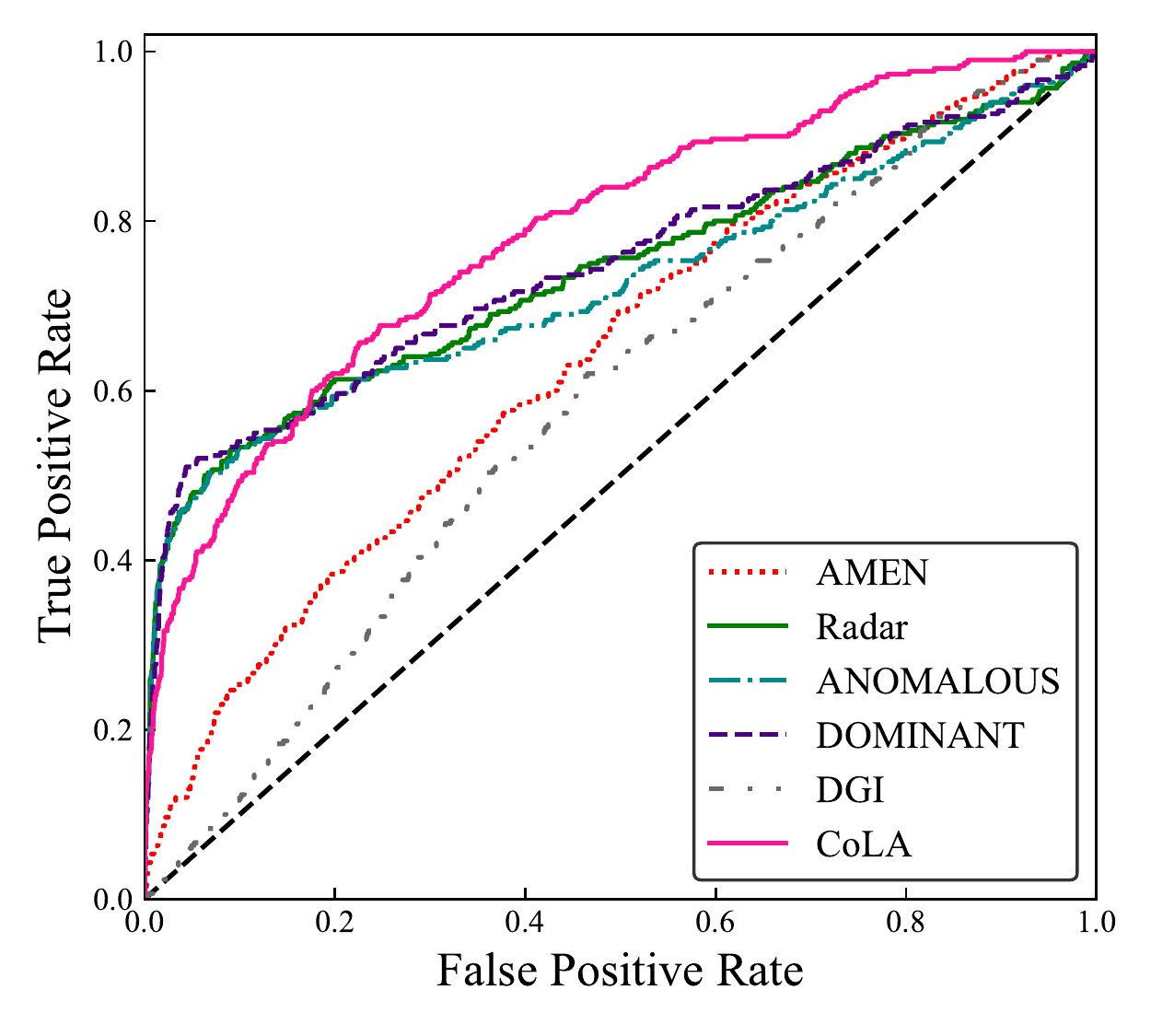}
	}\hspace{-4mm}
	\subfigure[Flickr]{
		\includegraphics[width=4.5cm]{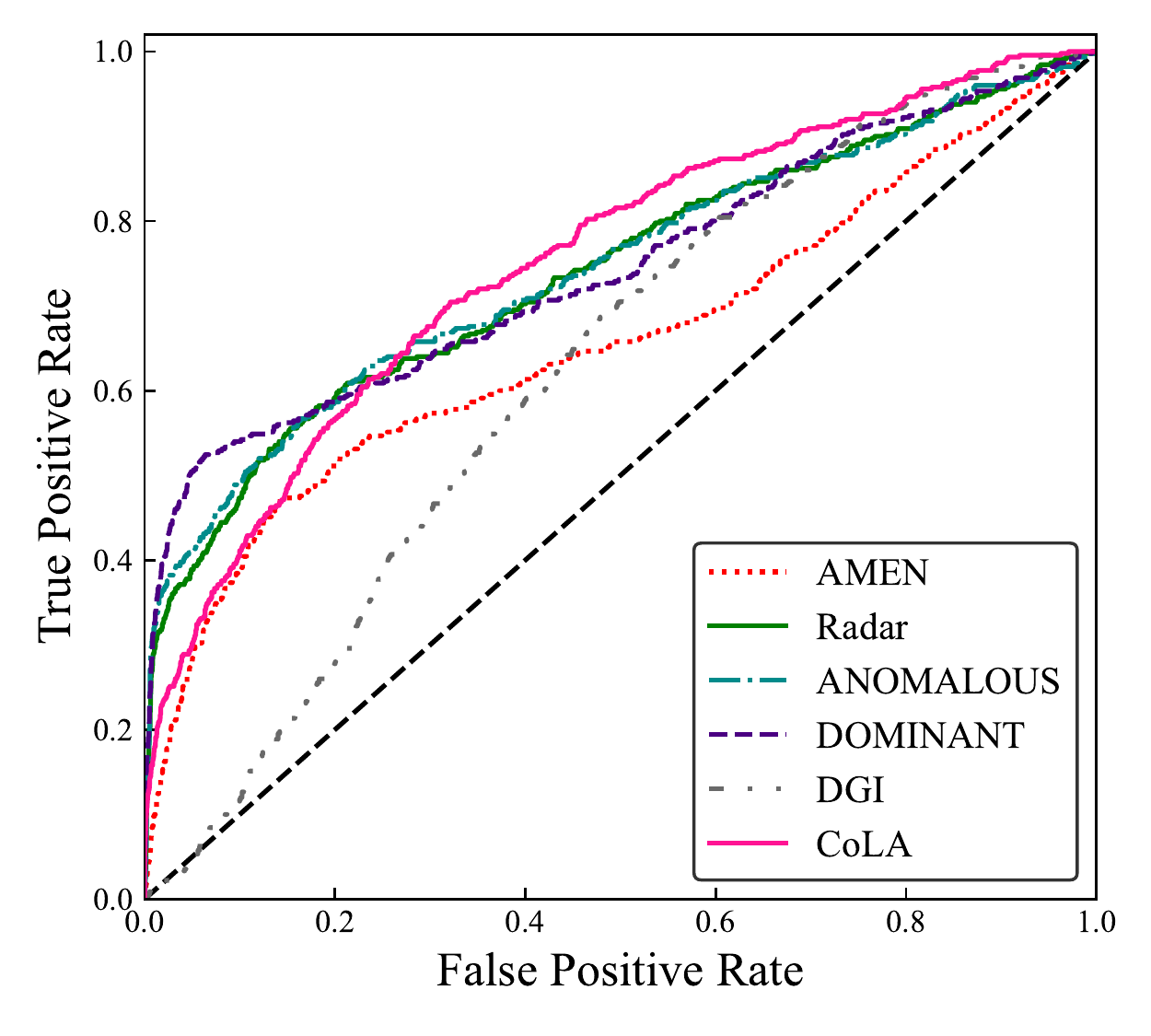}
	}\hspace{-4mm}
	\subfigure[ACM]{
		\includegraphics[width=4.5cm]{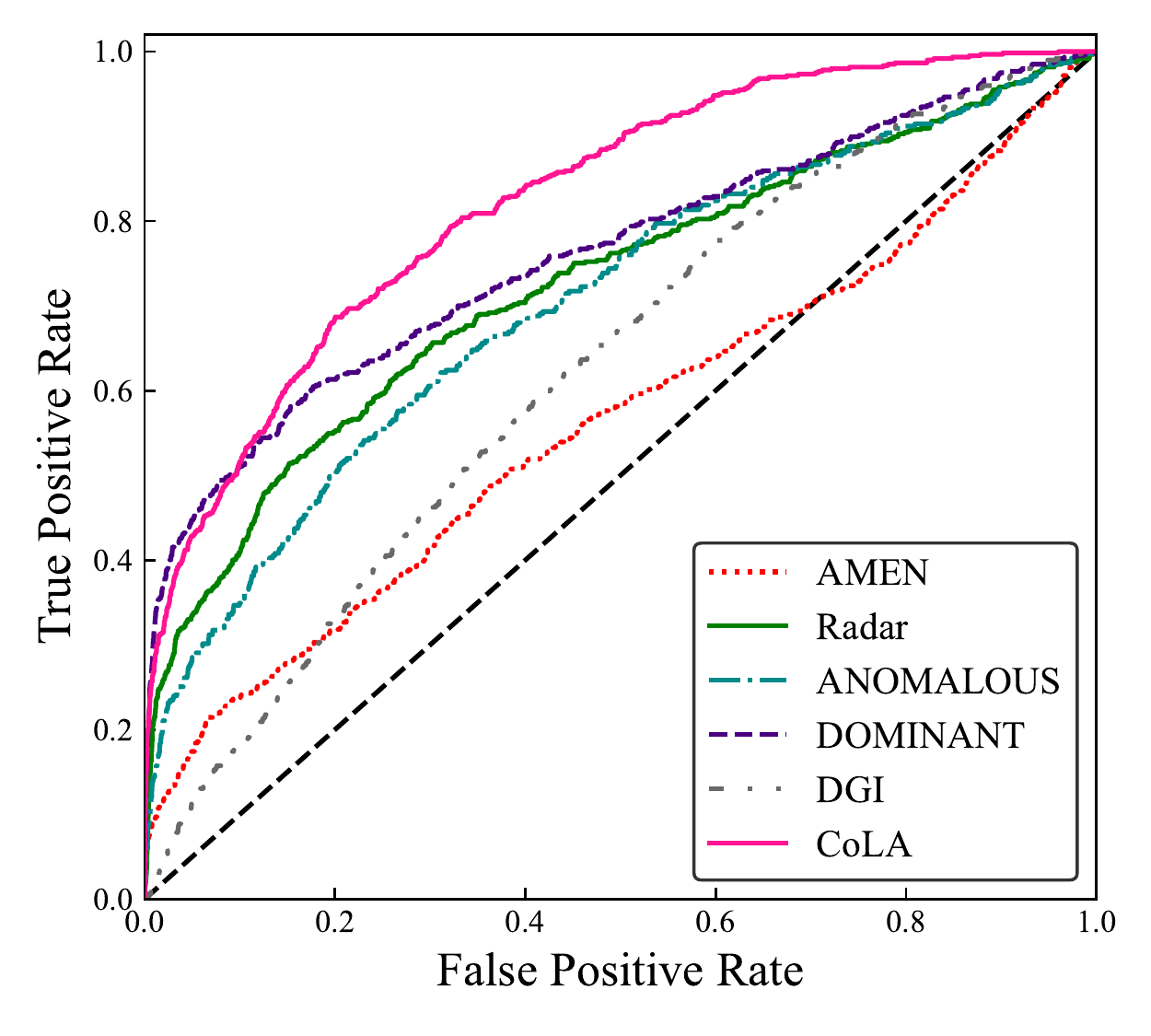}
	}\hspace{-4mm}
	\subfigure[Cora]{
		\includegraphics[width=4.5cm]{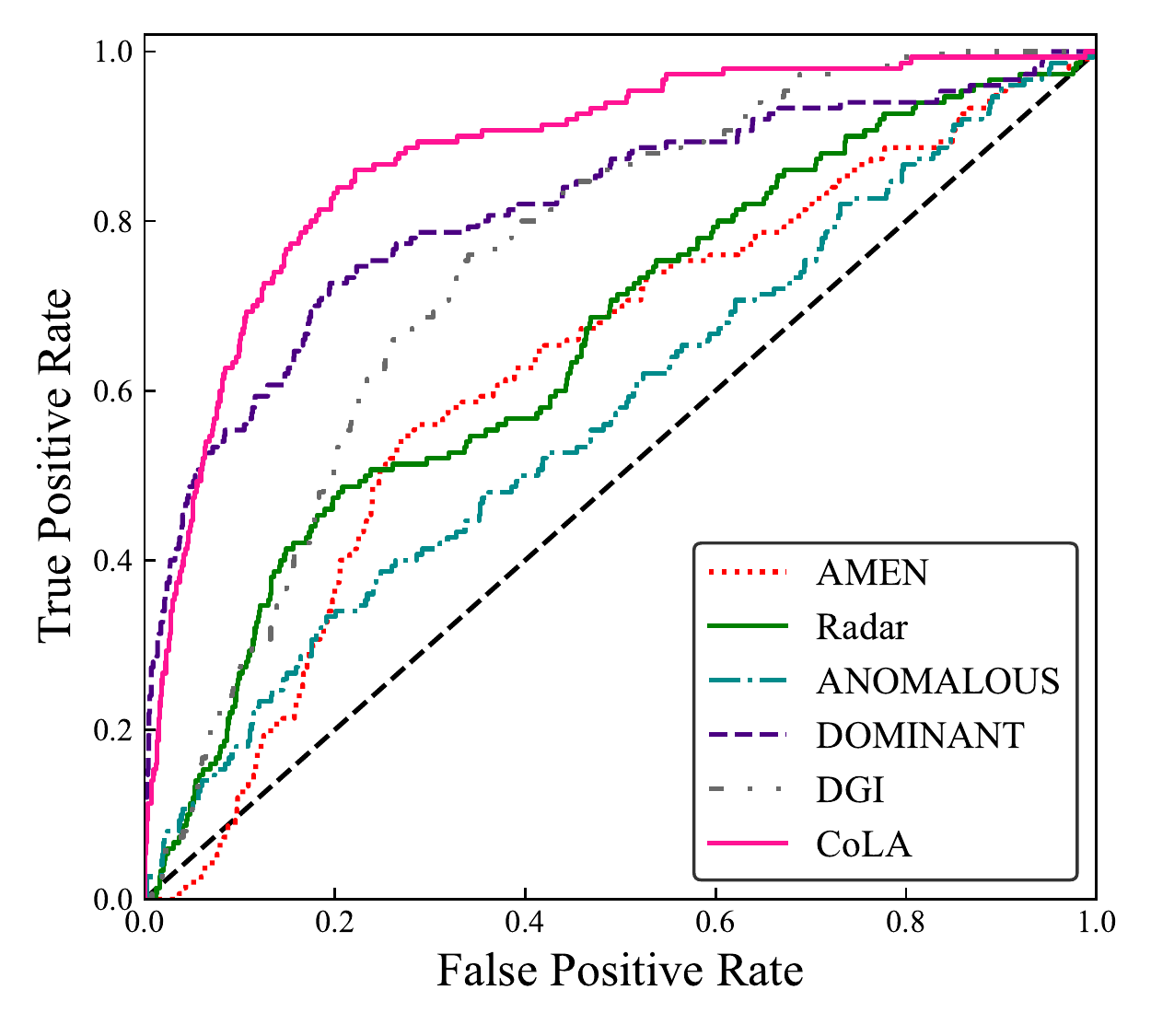}
	}
	\subfigure[Citeseer]{
		\includegraphics[width=4.5cm]{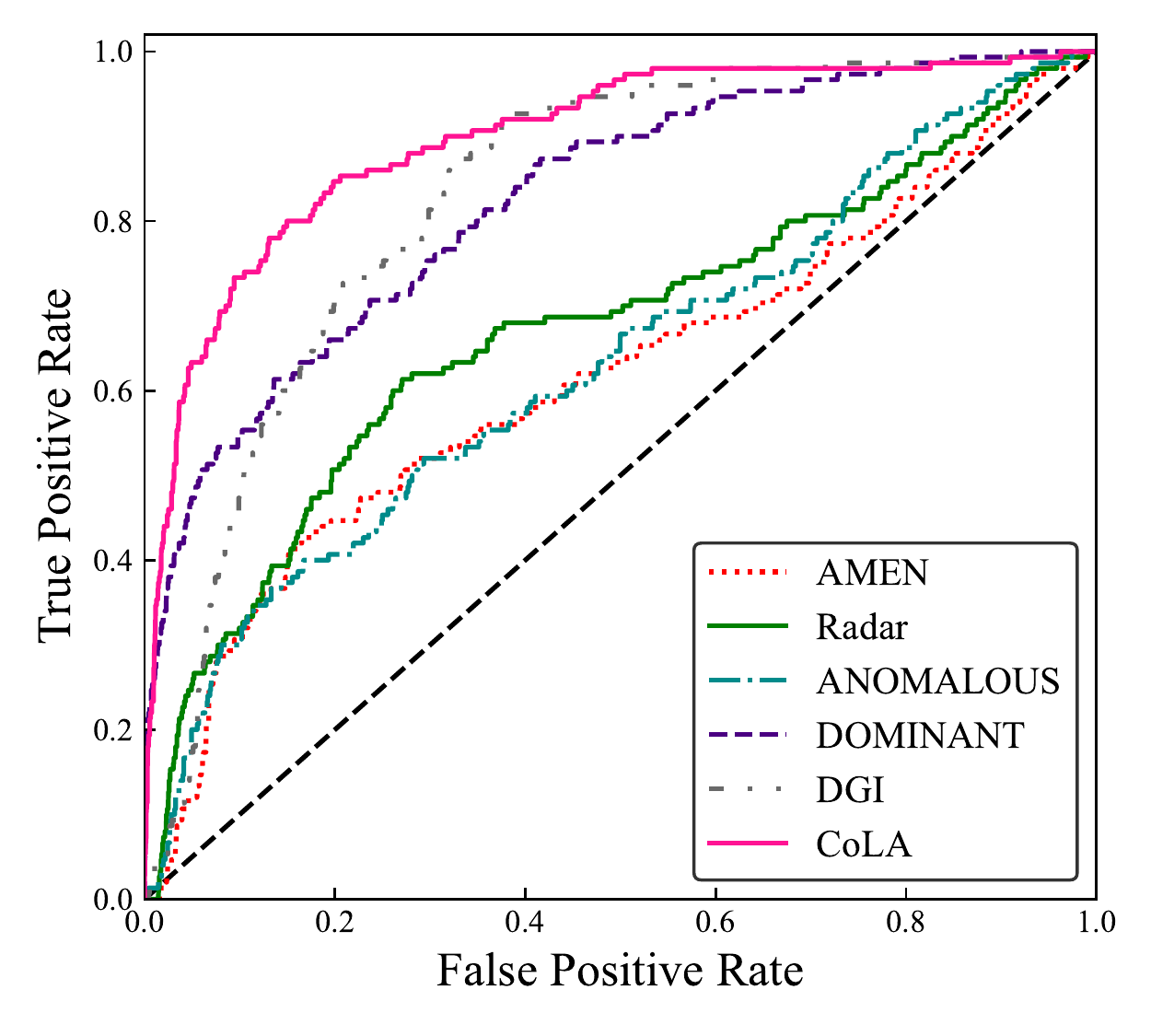}
	}\hspace{-4mm}
	\subfigure[Pubmed]{
		\includegraphics[width=4.5cm]{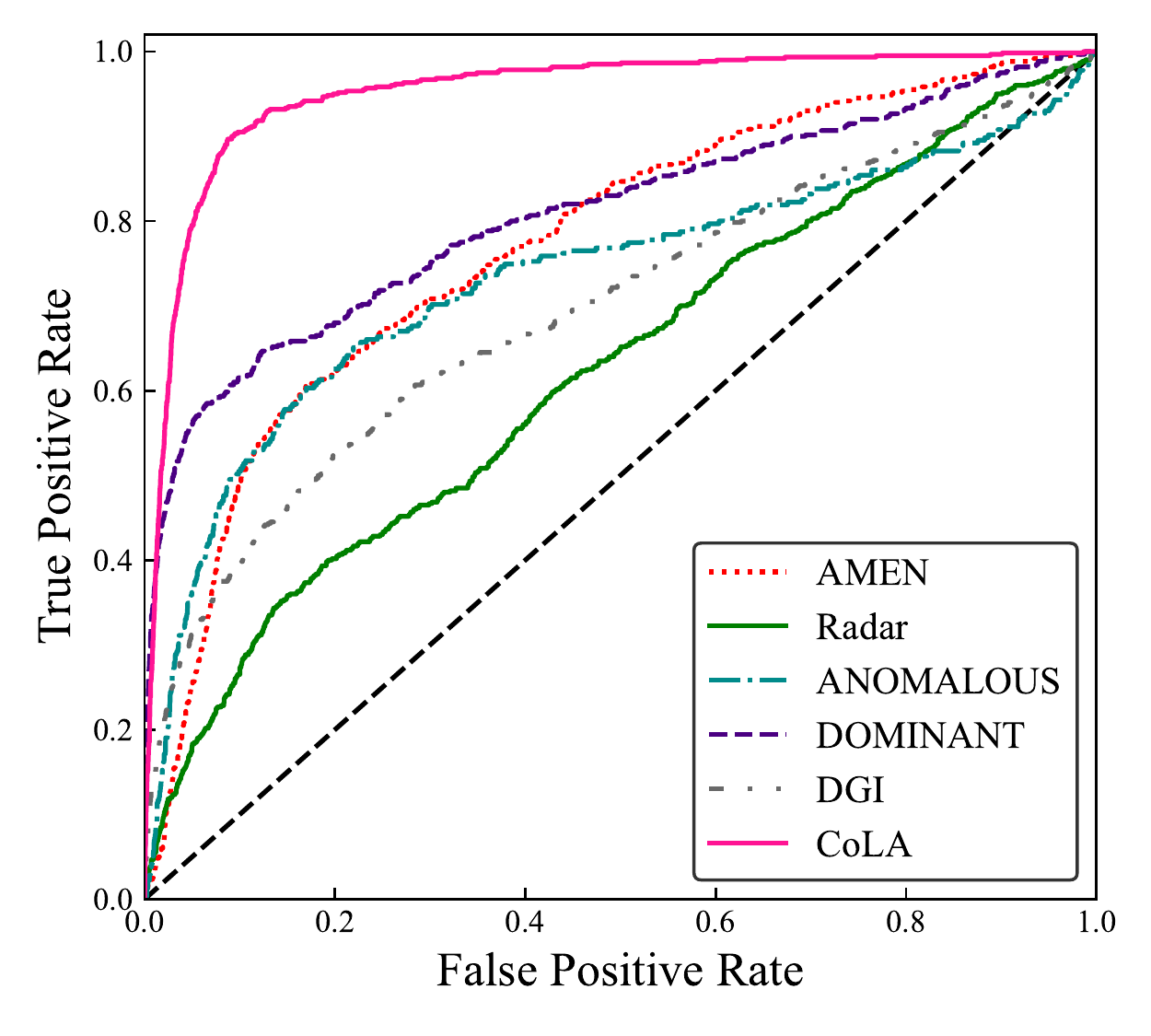}
	}\hspace{-4mm}
	\subfigure[ogbn-arxiv]{
	\includegraphics[width=4.5cm]{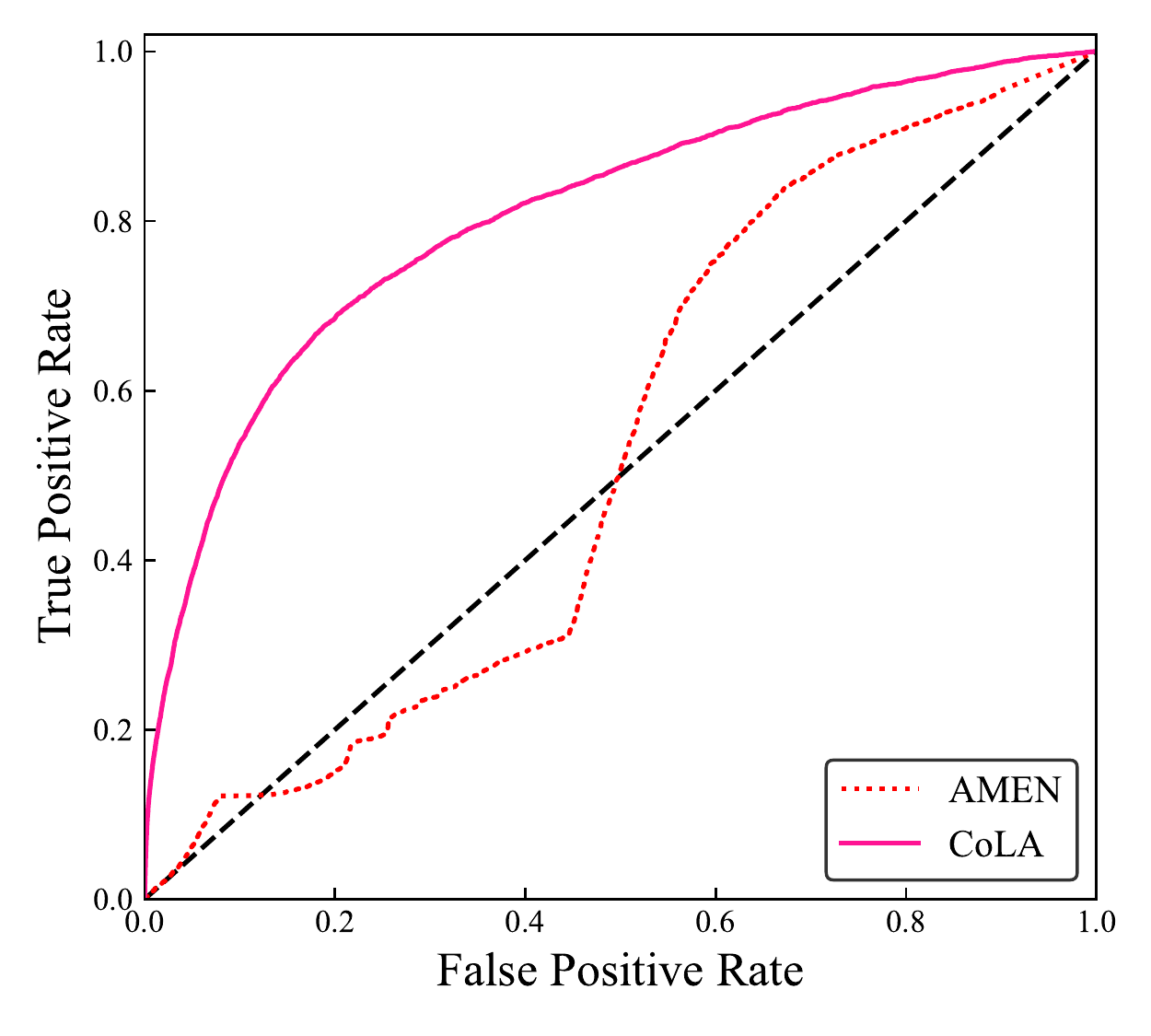}
	}
	\caption{ROC curves comparison on seven benchmark datasets. The area under the curve is larger, the anomaly detection performance is better. The dashed line is the ``random line'' which indicates the performance under  randomly guessing.}
	\label{fig:ROC}
\end{figure*}

\begin{table*}[!htbp]
	\small
	\centering
	\caption{AUC values comparison on seven benchmark datasets. OOM means the issue Out-Of-Memory is incurred. The best performing method in each experiment is in bold.}
	\begin{tabular}{ p{70 pt}<{\centering}|p{40 pt}<{\centering}p{40 pt}<{\centering}p{40 pt}<{\centering}p{40 pt}<{\centering}p{40 pt}<{\centering}p{40 pt}<{\centering}p{40 pt}<{\centering}}   
		\toprule[1.0pt]
		Methods & {Blogcatalog} & {Flickr} & {ACM} & {Cora}  & {Citeseer} & {Pubmed} & {ogbn-arxiv} \\
		\cmidrule{1-8}
		AMEN            & 0.6392 & 0.6573 & 0.5626 & 0.6266 & 0.6154 & 0.7713 & 0.5279\\		
		Radar             & 0.7401 & 0.7399 & 0.7247 & 0.6587 & 0.6709 & 0.6233 & OOM\\
		ANOMALOUS & 0.7237 & 0.7434 & 0.7038 & 0.5770 & 0.6307 & 0.7316 & OOM \\
		DOMINANT     & 0.7468 & 0.7442 & 0.7601 & 0.8155 & 0.8251 & 0.8081 & OOM\\
		DGI     & 0.5827 & 0.6237 & 0.6240 & 0.7511 & 0.8293 & 0.6962 & OOM\\
		\cmidrule{1-8}
		CoLA              & \textbf{0.7854} & \textbf{0.7513} & \textbf{0.8237} & \textbf{0.8779} & \textbf{0.8968} & \textbf{0.9512} & \textbf{0.8073} \\
		\bottomrule[1.0pt]
	\end{tabular}
	\label{table:AUC}
\end{table*}

\subsubsection{Evaluation metrics}
To measure the performance of our proposed framework and the baselines, we employ ROC-AUC as the metrics. ROC-AUC is widely used in previous works for the evaluation of anomaly detection performance \cite{dominant_ding2019deep, radar_li2017radar, anomalous_peng2018anomalous}. The ROC curve is a plot of true positive rate (an anomaly is recognized as an anomaly) against false positive rate (a normal node is recognized as an anomaly) according to the ground-truth anomalous labels and the anomaly detection results. AUC value is the area under the ROC curve, which represents the probability that a randomly chosen abnormal node is ranked higher than a normal node. The AUC which is close to $1$ means the method has high performance.

\subsubsection{Parameter Settings}
For the sake of efficiency and performance, we fixed the size $S$ of the sampled subgraph (number of nodes in the subgraph) to $4$. For isolated nodes or the nodes which belong to a community with a size smaller than the predetermined subgraph size, we sample the available nodes repeatedly until an overlapping subgraph with the set size is obtained. In the GNN-based contrastive learning model, the layer number of the GNN module is set to $1$ since it is enough to extract the information of subgraphs with a small size. The embedding dimension is fixed to be $64$. In the training phase, the batch size $B$ is set to $300$ for each dataset. Adam \cite{adam_kingma2014adam} optimization algorithm is employed to train the contrastive learning model. We train the model for BlogCatalog, Flickr and ACM datasets with 400 epochs, and train on Cora, Citeseer and Pubmed datasets with 100 epochs. The learning rates for Cora, Citeseer, Pubmed and Flickr are $0.001$, while the learning rates for BlogCatalog and ACM are set to $0.003$ and $0.0005$, respectively. For ogbn-arxiv dataset, we train for $2,000$ epochs using a learning rates of $0.0001$. In the inference phase, we sample $256$ rounds to acquire accurate detection results for each dataset. We run our proposed framework for 10 times and report the average results to prevent extreme cases. For the consideration of detection performance and efficiency, we use PCA \cite{pca_wold1987principal} to reduce the dimension of attributes to $30$ before we run the shallow baselines (AMEN, Radar and ANOMALOUS). For DGI We employ Equation (\ref{eq:ano_func}) to compute the anomaly score.

\subsubsection{Computing Infrastructures}
Our proposed learning framework is implemented using PyTorch 1.4.0 \cite{paszke2019pytorch}. For RWR subgraph sampling, we use the existing graph sampling function from library DGL 0.3.1 \cite{wang2019dgl}. The computation of ROC and AUC is acquired by Scikit-learn \cite{pedregosa2011scikit}. All experiments are conducted on a personal computer with Ubuntu 16.04 OS, an NVIDIA GeForce RTX 2070 (8GB memory) GPU, an Intel Core i7-7700k (4.20 GHz) CPU and 15.6 GB of RAM.

\subsection{Anomaly Detection Results} \label{subsec:exp_ad_result}

In this subsection, we evaluate the anomaly detection performance of the proposed framework by comparing it with the baseline methods. The comparison of ROC curves is demonstrated in Figure \ref{fig:ROC}. By calculating the area under the ROC curves, the AUC scores of the seven benchmark datasets are shown in Table \ref{table:AUC} for comparison. 
According to the results, we have the following observations:

\begin{figure*}[ht]
	\centering
	\subfigure[Sampling rounds w.r.t. AUC values]{
		\includegraphics[width=5.7cm]{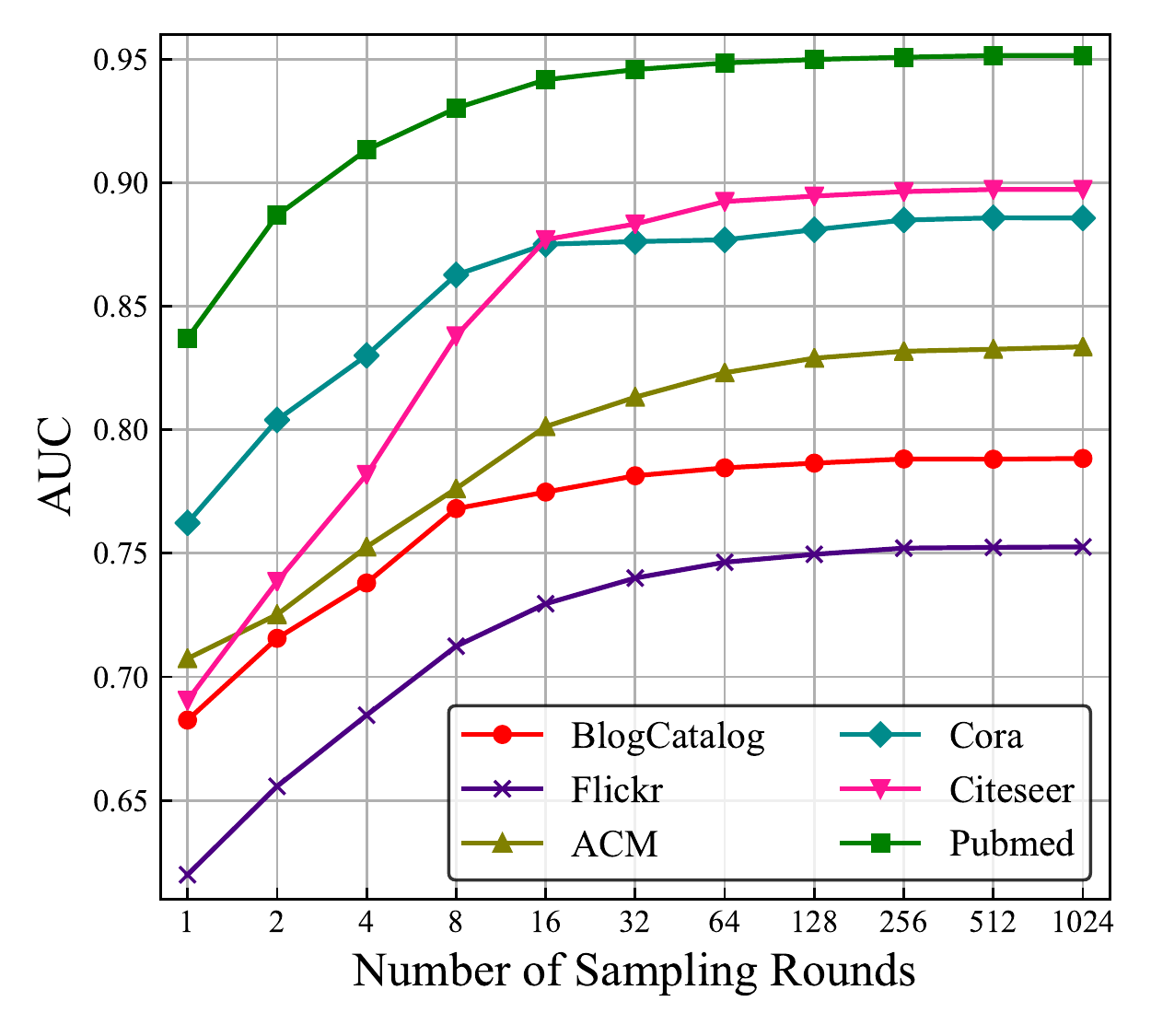}
		\label{fig:parameter_round}
	}
	\subfigure[Subgraph sizes w.r.t. AUC values]{
		\includegraphics[width=5.7cm]{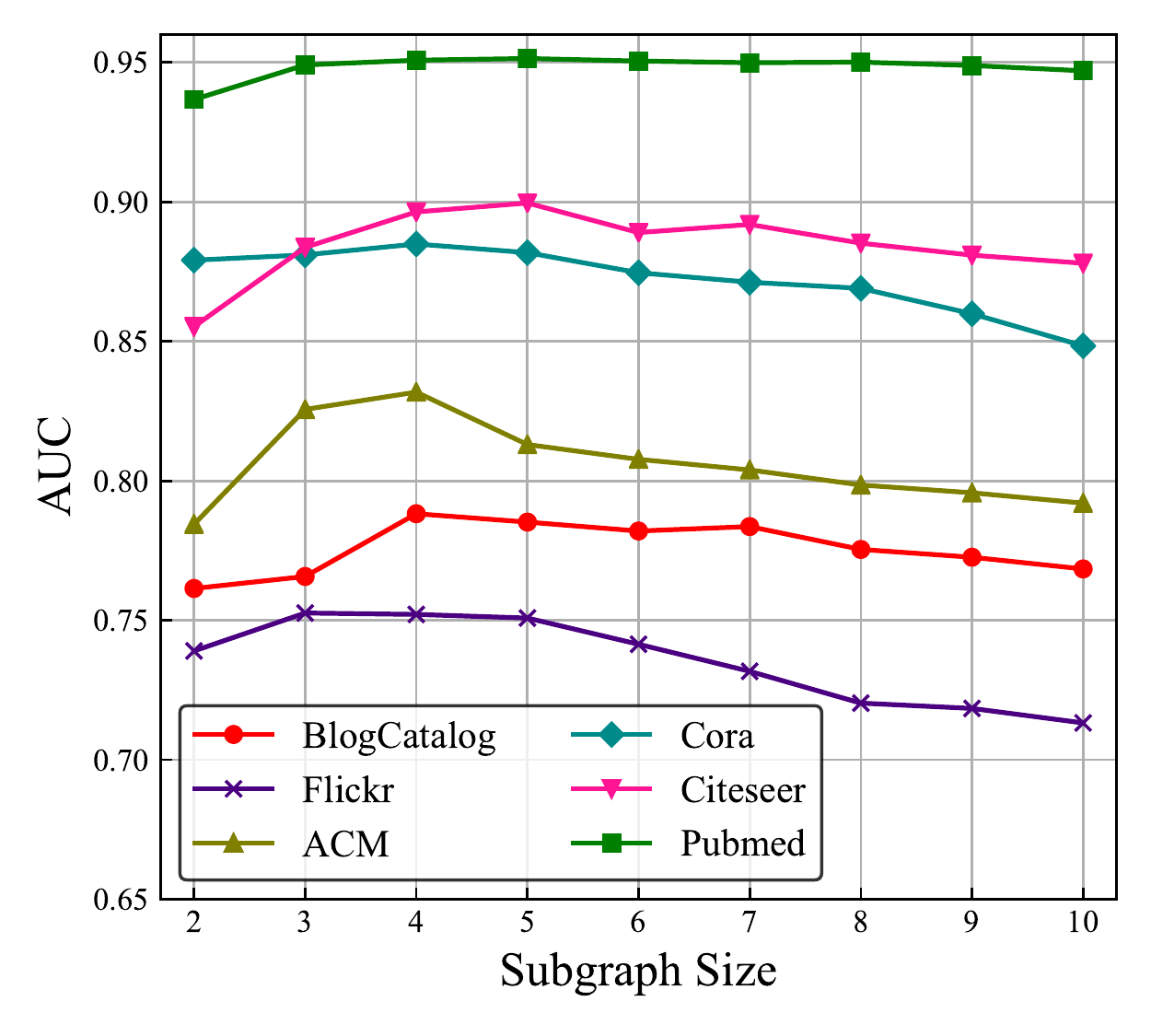}
		\label{fig:parameter_subgraph_size}
	}
	\subfigure[Embedding dimension w.r.t. AUC values]{
		\includegraphics[width=5.7cm]{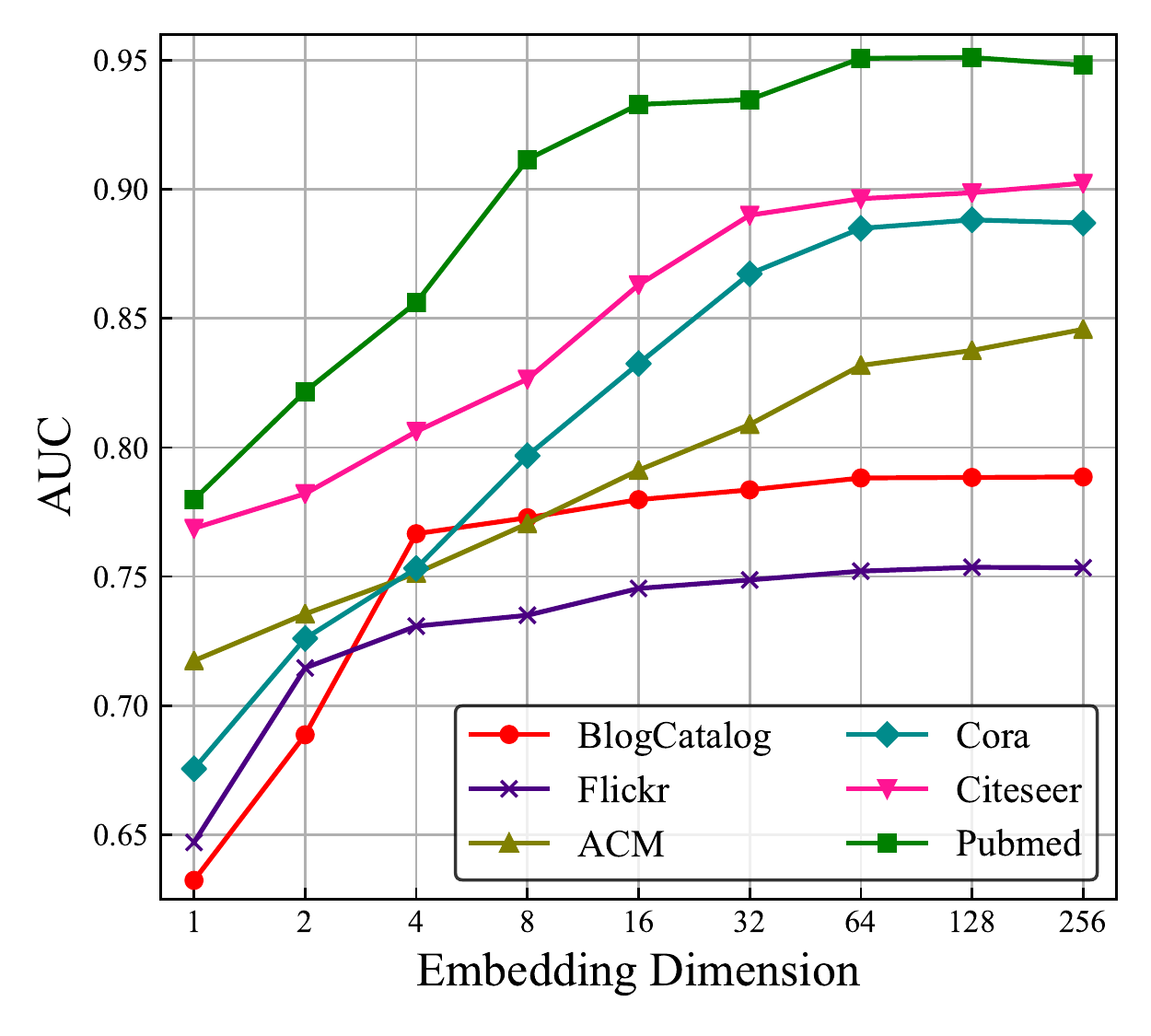}
		\label{fig:parameter_embedding_dim}
	}
	\caption{The experimental results for parameter study. The subfigure (a), (b) and (c) shows the impact of different sampling rounds, subgraph sizes and embedding dimension w.r.t. AUC values respectively. }
	\label{fig:para_study}
\end{figure*}

\begin{itemize} 
	\item On all seven datasets, our proposed CoLA achieves the best anomaly detection performance. In particular, compared with the best results of the baselines, our framework obtains a significant improvement of $6.44\%$ on AUC averagely. The main reason is that CoLA successfully captures the relationship between each node and its local substructure with the instance pair sampling, and extracts discriminative scores from the contextual and structural information with the GNN-based contrastive learning model.
	
	\item Compared to the deep learning-based methods, the shallow methods, AMEN, Radar and ANOMALOUS, cannot achieve satisfying results. Their performance is limited by the shallow mechanisms to deal with the high-dimensional node attributes and the sparse, complex network structures. 
	
	\item The contrastive learning method, DGI, does not show competitive performance, even if it uses a contrastive mechanism and anomaly score function similar to CoLA. The reason is that it adopts ``full graph v.s. node'' instance pair when performing contrastive learning, which cannot capture the abnormality in local substructure. In the contrast, the ``target node v.s. local subgraph'' leveraged by CoLA is sensitive to the local abnormal information.
	
	\item Compared to the autoencoder-based deep method DOMINANT, CoLA achieves significant performance gains, especially on the five citation network datasets. Two main reasons are: (1). CoLA well exploits the network data by constructing the instance pairs, instead of simply reconstructing the original data; (2). The objective of CoLA is related to the target of anomaly detection, which can train the learning model to generate discriminative scores for the final abnormality ranking. 
	 
	\item CoLA has better performance advantages on the five citation network datasets (ACM, Cora, Citeseer, Pubmed and ogbn-arxiv). The possible reason is that the mean degrees of citation networks ($1.43$ to $6.89$) are much smaller than those of social networks ($31.64$ to $33.05$). Therefore, on citation networks, the sampled substructure for each node has a better consistency under multiply rounds of sampling, which makes the model can capture the abnormality of each node more clearly. 
	
	\item CoLA is successfully run on the large-scale network dataset ogbn-arxiv, while most of the baselines (Radar, ANOMALOUS, DOMINANT, and DGI) fail to output the detection results due to their large requirement of memory. Meanwhile, our framework also outperforms the shallow method AMEN by a wide margin. The reason why CoLA can detect anomaly on large-scale networks is that the space complexity of CoLA is independent of the number of nodes $n$, which has been analyzed in Section \ref{subsec:model_framework}.
	
\end{itemize}

\subsection{Parameter Study} \label{subsec:exp_param_study}  

In this subsection, we investigate the impacts of three important parameters on the performance of the proposed framework: the number of sampling rounds, the size of subgraph, and the dimension of latent embedding. We only perform these experiments on the six small-scale datasets owing to the limitation of efficiency. 

\subsubsection{Effect of the number of sampling rounds $R$}
In this experiment, we modify the value of $R$ to study its impact on AUC. The performance variance results are demonstrated in Figure \ref{fig:parameter_round}. As we can see, when the detection results are only computed with one-shot sampling, the detection performance is poor. With the sampling rounds growing, there is a significant boost in the AUC of each dataset within a certain range. However, when $R$ is larger than $256$, the performance improvement obtained by setting a larger $R$ becomes little. The experiment results empirically prove our analysis in Subsection \ref{subsec:model_anoscore}: with $R$ gets larger, the estimation for abnormalities for each node becomes more accurate. On the basis of this result, we set $R=256$ in other experiments to balance performance and efficiency.

\subsubsection{Effect of subgraph size $c$}
We further analyze the significance of the subgraph size $c$ on different datasets. We report the AUC scores over different choices of subgraph sizes in Figure \ref{fig:parameter_subgraph_size}. As shown in the figure, when $c$ is extremely small ($c=2$), the AUC is relatively low. The possible reason is that, in such a situation, only the target node itself and one of its neighbors are concluded in the subgraph, but there is no structural information other than the connection between these two nodes is considered. The lack of enough neighboring structural information results in poor performance. Within certain value ranges, the AUC increases follow $c$. Then, when $c>5$, the detection performance declines with $c$ get larger. A reasonable description is that: when the subgraph size is large, the nodes with relatively long distance to the target node will be sampled into the subgraph. However, these remote nodes are generally independent of the abnormality, which becomes the ``noise information'' for our detection task. For all datasets, better detection performance can be obtained when the value of $S$ is around $4$. Consequently, we fix the value of $c$ to $4$ for the sake of running efficiency and robustness over all datasets.

\subsubsection{Effect of embedding dimension $d$}
We explore the sensitivity of embedding dimension $d$ for CoLA framework. We alter the value of $d$ to see how it affects the performance of our method. The performance change of CoLA is illustrated in Figure \ref{fig:parameter_embedding_dim}. For each dataset, the AUC value increases with the embedding dimension growing. When adding the dimension of embedding from $1$-neuron to $32$-neurons, the performance of anomaly detection steadily rises; but when we further increase $d$, the performance gain becomes light. We observed that, for most of the datasets, $64$-dimension latent embedding can provide sufficient information for the downstream contrastive learning and anomaly detection tasks. As a result, we set the hyper-parameter $d$ to $64$ for efficiency consideration.

\subsection{Ablation Study} \label{subsec:exp_ablation_study}    
\begin{table*}[!htbp]
	\small
	\centering
	\caption{Effect of different readout function on AUC values. The best performing method in each experiment is in bold.}
	\begin{tabular}{ p{100 pt}<{\centering}|p{40 pt}<{\centering}p{40 pt}<{\centering}p{40 pt}<{\centering}p{40 pt}<{\centering}p{40 pt}<{\centering}p{40 pt}<{\centering}}    
		\toprule[1.0pt]
		  & {BlogCatalog} & {Flickr} & {ACM} & {Cora}  & {Citeseer} & {Pubmed}  \\
		\cmidrule{1-7}
		Max Pooling                      & 0.7787 & 0.7507 & \textbf{0.8303} & 0.8681 & 0.8849 & 0.9508 \\		
		Min Pooling                      & 0.7567 & 0.7389 & 0.7968 & 0.8553 & 0.8743 & 0.9350 \\
		Weighted Average Pooling         & 0.7801 & 0.7494 & 0.8175 & 0.8764 & 0.8927 & \textbf{0.9523} \\		
		\cmidrule{1-7}
		CoLA(Average Pooling)                   & \textbf{0.7854} & \textbf{0.7513} & 0.8237 & \textbf{0.8779} & \textbf{0.8968} & 0.9512 \\
		\bottomrule[1.0pt]
	\end{tabular}
	\label{table:ablation_readout}
\end{table*}

\begin{table*}[!htbp]
	\small
	\centering
	\caption{Effect of different source of score computation on AUC values. The best performing method in each experiment is in bold.}
	\begin{tabular}{ p{100 pt}<{\centering}|p{40 pt}<{\centering}p{40 pt}<{\centering}p{40 pt}<{\centering}p{40 pt}<{\centering}p{40 pt}<{\centering}p{40 pt}<{\centering}}  
		\toprule[1.0pt]
		  & {BlogCatalog} & {Flickr} & {ACM} & {Cora}  & {Citeseer} & {Pubmed}  \\
		\cmidrule{1-7}
		CoLA(+)                      & 0.7551 & 0.7213 & 0.8002 & 0.8658 & 0.8571 & 0.9509 \\		
		CoLA(-)                      & 0.7745 & 0.7502 & 0.7718 & 0.6891 & 0.8254 & 0.6531 \\
		\cmidrule{1-7}
		CoLA(+/-)                   & \textbf{0.7854} & \textbf{0.7513} & \textbf{0.8237} & \textbf{0.8779} & \textbf{0.8968} & \textbf{0.9512} \\
		\bottomrule[1.0pt]
	\end{tabular}
	\label{table:ablation_source}
\end{table*}

\begin{table*}[!htbp]
	\small
	\centering
	\caption{Effect of different score estimation mode on AUC values. The best performing method in each experiment is in bold.}
	\begin{tabular}{ p{100 pt}<{\centering}|p{40 pt}<{\centering}p{40 pt}<{\centering}p{40 pt}<{\centering}p{40 pt}<{\centering}p{40 pt}<{\centering}p{40 pt}<{\centering}}    
		\toprule[1.0pt]
		  & {BlogCatalog} & {Flickr} & {ACM} & {Cora}  & {Citeseer} & {Pubmed}  \\
		\cmidrule{1-7}
		CoLA(min)                  & 0.7611 & 0.7219 & 0.5407 & 0.7525 & 0.6279 & 0.7987 \\		
		CoLA(mean+min)             & 0.7729 & 0.7371 & 0.6903 & 0.8165 & 0.7599 & 0.8796 \\
		CoLA(max)                  & 0.6465 & 0.6071 & 0.7093 & 0.8152 & 0.7906 & 0.8839 \\
		CoLA(mean+max)        & 0.7383 & 0.6989 & 0.7652 & 0.8678 & 0.8638 & 0.9359 \\
		CoLA(std)                    & 0.3719 & 0.3581 & 0.8037 & 0.5453 & 0.7515 & 0.7035 \\
		CoLA(mean+std)          & 0.7665 & 0.7241 & \textbf{0.8372} & \textbf{0.8869} & \textbf{0.9047} &\textbf{0.9532} \\
		CoLA(-std)                   & 0.6316 & 0.6449 & 0.2149 & 0.4256 & 0.2455 & 0.2969 \\
		CoLA(mean-std)           & \textbf{0.7910} & \textbf{0.7526} & 0.7562 & 0.8479 & 0.8608 & 0.9387 \\
		\cmidrule{1-7}
		CoLA(mean)       & 0.7854 &  0.7513  & 0.8237 & 0.8779 & 0.8968 & 0.9512 \\
		\bottomrule[1.0pt]
	\end{tabular}
	\label{table:ablation_mode}
\end{table*}

In this subsection, we study the effect of changing different readout functions, the source of anomaly score computation, and the estimation mode of anomaly score. Here we only discuss the performance on the six small-scale datasets.

\subsubsection{Effect of readout function}
In this experiment, we investigate the choice of different types of readout functions in our framework. We carry out the experiments on three possible readout functions: Max Pooling, Min Pooling, and Weighted Average Pooling. Max/Min Pooling is to collect the maximum/minimum value on each dimension to generate the pooled vector. Weighted Average Pooling first takes the embedding of the target node as a ``query'' and calculates the similarities between query and node embeddings in the local subgraph by inner production. Then, a Softmax function is utilized to regularize the similarities which further serve as ``weights'' to compute the readout output with a weighted average. Note that CoLA adopts Average Pooling, which is simpler than Weighted Average Pooling. 

The experimental results are illustrated in Table \ref{table:ablation_readout}. Compared to other readout function, Min Pooling always has a minor performance, which means that using the minimum value would lead to lose of information. 
Max Pooling has competitive performance on most of the datasets, but is not the best. Although Weighted Average Pooling costs heavier computation, it has a close performance to Max Pooling, which indicates that the similarity-based weighted average may lead to a sub-optimal solution for subgraph readout. Compared to other readout functions, CoLA with Average Pooling achieves the best results on 4 of 6 datasets. On ACM and Pubmed, it also shows competitive performance, which evinces that Average Pooling has a better capability of generalization.

\subsubsection{Effect of the source of score computation}
In Equation (\ref{eq:ano_func}), we consider the predicted scores of both positive instances and negative instances as the source of anomaly score computation. Here, we investigate the contribution of each term. Table \ref{table:ablation_mode} shows the AUC values of two variants of our framework: CoLA(+) denotes that only the predicted scores of positive instances are used to calculate the anomaly scores, and CoLA(-) means that only the negative scores are considered. Finally, CoLA(+/-) is the full version that considers both sides. As we can observe, for two social network datasets, the predicted scores of negative instances are more helpful to the final result. Nevertheless, for the remaining citation network datasets, the positive instances have a greater contribution. Despite the above difference, considering both positive and negative scores is always beneficial to anomaly detection performance.

\subsubsection{Effect of estimation mode of anomaly score}
In CoLA, we compute the mean values as the anomaly scores, which is a standard estimation mode for multiple sampling. In this experiment, we discuss the effect of other estimation modes. As shown in Table \ref{table:ablation_mode}, we carry out our experiment on six estimation modes: CoLA(max)/CoLA(min) means using the maximum/minimum value of multi-round predictions as anomaly scores; CoLA(std) and CoLA(-std) adopt the standard deviation and the opposite of standard deviation as anomaly score, respectively; the rest three situations consider the sum of mean value and corresponding terms as an estimation. 

We make the following observations: 

\begin{itemize} 
	
	\item Using maximum/minimum value as the anomaly score is less effective than using mean value. Considering both the maximum/minimum and the mean value can obtain better performance, but it is still worse than only using the mean value.
	
	\item Introducing the standard deviation can bring extra performance improvement. However, the correlation between abnormality and standard deviation varies with different datasets: for BlogCatalog and Flickr, there is a negative correlation between abnormality and standard deviation; on contrary, a positive correlation is shown for the citation networks. 
	
	\item Calculating the mean value is not the best but the most robust choice. For all datasets, we can obtain a relatively good detection performance with CoLA(mean). Furthermore, compared with only using maximum/minimum value or standard deviation, the introduction of additional mean values will lead to a better result.
	
\end{itemize}

\section{Conclusion}\label{sec:conclusion}

In this paper, we make the first attempt to apply contrastive self-supervised learning to the anomaly detection problem of attributed networks. We propose a novel anomaly detection framework, CoLA, which is consisted of three components: contrastive instance pair sampling, GNN-based contrastive learning model, and multi-round sampling-based anomaly score computation. Our model successfully captures the relationship between each node and its neighboring structure and uses an anomaly-related objective to train the contrastive learning model. A series of experiments on seven benchmark datasets demonstrate the effectiveness and superiority of the proposed framework in solving the anomaly detection problems on attributed networks.

We believe that the proposed framework opens a new opportunity to expand self-supervised learning and contrastive learning to increasingly graph anomaly detection applications. In future works, we will extend self-supervised contrastive learning-based anomaly detection methods to more complex network/graph data, e.g., heterogeneous graph, spatial-temporal graph, and dynamic graph.

\ifCLASSOPTIONcaptionsoff
  \newpage
\fi

\bibliography{tnnls}

\begin{thebibliography}{10}
\providecommand{\url}[1]{#1}
\csname url@samestyle\endcsname
\providecommand{\newblock}{\relax}
\providecommand{\bibinfo}[2]{#2}
\providecommand{\BIBentrySTDinterwordspacing}{\spaceskip=0pt\relax}
\providecommand{\BIBentryALTinterwordstretchfactor}{4}
\providecommand{\BIBentryALTinterwordspacing}{\spaceskip=\fontdimen2\font plus
\BIBentryALTinterwordstretchfactor\fontdimen3\font minus
  \fontdimen4\font\relax}
\providecommand{\BIBforeignlanguage}[2]{{%
\expandafter\ifx\csname l@#1\endcsname\relax
\typeout{** WARNING: IEEEtran.bst: No hyphenation pattern has been}%
\typeout{** loaded for the language `#1'. Using the pattern for}%
\typeout{** the default language instead.}%
\else
\language=\csname l@#1\endcsname
\fi
#2}}
\providecommand{\BIBdecl}{\relax}
\BIBdecl

\bibitem{liu2018heterogeneous}
Z.~Liu, C.~Chen, X.~Yang, J.~Zhou, X.~Li, and L.~Song, ``Heterogeneous graph
  neural networks for malicious account detection,'' in \emph{Proceedings of
  the 27th ACM International Conference on Information and Knowledge
  Management}, 2018, pp. 2077--2085.

\bibitem{tang2009relational}
L.~Tang and H.~Liu, ``Relational learning via latent social dimensions,'' in
  \emph{Proceedings of the 15th ACM SIGKDD international conference on
  Knowledge discovery and data mining}, 2009, pp. 817--826.

\bibitem{zhang2019your}
Y.~Zhang, Y.~Fan, W.~Song, S.~Hou, Y.~Ye, X.~Li, L.~Zhao, C.~Shi, J.~Wang, and
  Q.~Xiong, ``Your style your identity: Leveraging writing and photography
  styles for drug trafficker identification in darknet markets over attributed
  heterogeneous information network,'' in \emph{The World Wide Web Conference},
  2019, pp. 3448--3454.

\bibitem{ying2018graph}
R.~Ying, R.~He, K.~Chen, P.~Eksombatchai, W.~L. Hamilton, and J.~Leskovec,
  ``Graph convolutional neural networks for web-scale recommender systems,'' in
  \emph{Proceedings of the 24th ACM SIGKDD International Conference on
  Knowledge Discovery \& Data Mining}, 2018, pp. 974--983.

\bibitem{fan2019graph}
W.~Fan, Y.~Ma, Q.~Li, Y.~He, E.~Zhao, J.~Tang, and D.~Yin, ``Graph neural
  networks for social recommendation,'' in \emph{The World Wide Web
  Conference}, 2019, pp. 417--426.

\bibitem{gcn_kipf2017semi}
T.~N. Kipf and M.~Welling, ``Semi-supervised classification with graph
  convolutional networks,'' in \emph{International Conference on Learning
  Representations}, 2017.

\bibitem{gat_ve2018graph}
P.~Veličković, G.~Cucurull, A.~Casanova, A.~Romero, P.~Liò, and Y.~Bengio,
  ``Graph attention networks,'' in \emph{International Conference on Learning
  Representations}, 2018.

\bibitem{gin_xu2019powerful}
K.~Xu, W.~Hu, J.~Leskovec, and S.~Jegelka, ``How powerful are graph neural
  networks?'' in \emph{International Conference on Learning Representations},
  2019.

\bibitem{ying2018hierarchical}
Z.~Ying, J.~You, C.~Morris, X.~Ren, W.~Hamilton, and J.~Leskovec,
  ``Hierarchical graph representation learning with differentiable pooling,''
  in \emph{Advances in neural information processing systems}, 2018, pp.
  4800--4810.

\bibitem{zhang2017weisfeiler}
M.~Zhang and Y.~Chen, ``Weisfeiler-lehman neural machine for link prediction,''
  in \emph{Proceedings of the 23rd ACM SIGKDD International Conference on
  Knowledge Discovery and Data Mining}, 2017, pp. 575--583.

\bibitem{kipf2016variational}
T.~N. Kipf and M.~Welling, ``Variational graph auto-encoders,'' \emph{arXiv
  preprint arXiv:1611.07308}, 2016.

\bibitem{pang2020deep}
G.~Pang, C.~Shen, L.~Cao, and A.~v.~d. Hengel, ``Deep learning for anomaly
  detection: A review,'' \emph{arXiv preprint arXiv:2007.02500}, 2020.

\bibitem{dominant_ding2019deep}
K.~Ding, J.~Li, R.~Bhanushali, and H.~Liu, ``Deep anomaly detection on
  attributed networks,'' in \emph{Proceedings of the 2019 SIAM International
  Conference on Data Mining}.\hskip 1em plus 0.5em minus 0.4em\relax SIAM,
  2019, pp. 594--602.

\bibitem{ocsvm_chen2001one}
Y.~Chen, X.~S. Zhou, and T.~S. Huang, ``One-class svm for learning in image
  retrieval,'' in \emph{Proceedings 2001 International Conference on Image
  Processing (Cat. No. 01CH37205)}, vol.~1.\hskip 1em plus 0.5em minus
  0.4em\relax IEEE, 2001, pp. 34--37.

\bibitem{scan_xu2007scan}
X.~Xu, N.~Yuruk, Z.~Feng, and T.~A. Schweiger, ``Scan: a structural clustering
  algorithm for networks,'' in \emph{Proceedings of the 13th ACM SIGKDD
  international conference on Knowledge discovery and data mining}, 2007, pp.
  824--833.

\bibitem{amen_perozzi2016scalable}
B.~Perozzi and L.~Akoglu, ``Scalable anomaly ranking of attributed
  neighborhoods,'' in \emph{Proceedings of the 2016 SIAM International
  Conference on Data Mining}.\hskip 1em plus 0.5em minus 0.4em\relax SIAM,
  2016, pp. 207--215.

\bibitem{radar_li2017radar}
J.~Li, H.~Dani, X.~Hu, and H.~Liu, ``Radar: Residual analysis for anomaly
  detection in attributed networks.'' in \emph{IJCAI}, 2017, pp. 2152--2158.

\bibitem{anomalous_peng2018anomalous}
Z.~Peng, M.~Luo, J.~Li, H.~Liu, and Q.~Zheng, ``Anomalous: A joint modeling
  approach for anomaly detection on attributed networks.'' in \emph{IJCAI},
  2018, pp. 3513--3519.

\bibitem{pang2020deep2}
G.~Pang, C.~Shen, H.~Jin, and A.~van~den Hengel, ``Deep weakly-supervised
  anomaly detection,'' 2020.

\bibitem{pang2019deep}
G.~Pang, C.~Shen, and A.~van~den Hengel, ``Deep anomaly detection with
  deviation networks,'' in \emph{Proceedings of the 25th ACM SIGKDD
  International Conference on Knowledge Discovery \& Data Mining}, 2019, pp.
  353--362.

\bibitem{ruff2018deep}
L.~Ruff, R.~Vandermeulen, N.~Goernitz, L.~Deecke, S.~A. Siddiqui, A.~Binder,
  E.~M{\"u}ller, and M.~Kloft, ``Deep one-class classification,'' in
  \emph{International conference on machine learning}, 2018, pp. 4393--4402.

\bibitem{specae_li2019specae}
Y.~Li, X.~Huang, J.~Li, M.~Du, and N.~Zou, ``Specae: Spectral autoencoder for
  anomaly detection in attributed networks,'' in \emph{Proceedings of the 28th
  ACM International Conference on Information and Knowledge Management}, 2019,
  pp. 2233--2236.

\bibitem{cpc_oord2018representation}
A.~v.~d. Oord, Y.~Li, and O.~Vinyals, ``Representation learning with
  contrastive predictive coding,'' \emph{arXiv preprint arXiv:1807.03748},
  2018.

\bibitem{simclr_chen2020simple}
T.~Chen, S.~Kornblith, M.~Norouzi, and G.~Hinton, ``A simple framework for
  contrastive learning of visual representations,'' \emph{arXiv preprint
  arXiv:2002.05709}, 2020.

\bibitem{dim_hjelm2018learning}
R.~D. Hjelm, A.~Fedorov, S.~Lavoie-Marchildon, K.~Grewal, P.~Bachman,
  A.~Trischler, and Y.~Bengio, ``Learning deep representations by mutual
  information estimation and maximization,'' in \emph{International Conference
  on Learning Representations}, 2018.

\bibitem{Liao2017Attributed}
L.~Liao, X.~He, H.~Zhang, and T.~S. Chua, ``Attributed social network
  embedding,'' in \emph{arXiv:1705.04969}, 2017.

\bibitem{Pan2016Tri}
S.~Pan, J.~Wu, X.~Zhu, C.~Zhang, and Y.~Wang, ``Tri-party deep network
  representation,'' in \emph{IJCAI}, 2016, pp. 1895--1901.

\bibitem{yu2019self}
W.~{Yu}, W.~{Cheng}, C.~{Aggarwal}, B.~{Zong}, H.~{Chen}, and W.~{Wang},
  ``Self-attentive attributed network embedding through adversarial learning,''
  in \emph{2019 IEEE International Conference on Data Mining (ICDM)}, 2019, pp.
  758--767.

\bibitem{dl_lecun2015deep}
Y.~LeCun, Y.~Bengio, and G.~Hinton, ``Deep learning,'' \emph{nature}, vol. 521,
  no. 7553, pp. 436--444, 2015.

\bibitem{gnn_survey_wu2020comprehensive}
Z.~{Wu}, S.~{Pan}, F.~{Chen}, G.~{Long}, C.~{Zhang}, and P.~S. {Yu}, ``A
  comprehensive survey on graph neural networks,'' \emph{IEEE Transactions on
  Neural Networks and Learning Systems}, pp. 1--21, 2020.

\bibitem{gori2005new}
M.~Gori, G.~Monfardini, and F.~Scarselli, ``A new model for learning in graph
  domains,'' in \emph{Proceedings. 2005 IEEE International Joint Conference on
  Neural Networks, 2005.}, vol.~2.\hskip 1em plus 0.5em minus 0.4em\relax IEEE,
  2005, pp. 729--734.

\bibitem{bruna2013spectral}
J.~Bruna, W.~Zaremba, A.~Szlam, and Y.~LeCun, ``Spectral networks and locally
  connected networks on graphs,'' \emph{arXiv preprint arXiv:1312.6203}, 2013.

\bibitem{henaff2015deep}
M.~Henaff, J.~Bruna, and Y.~LeCun, ``Deep convolutional networks on
  graph-structured data,'' \emph{arXiv preprint arXiv:1506.05163}, 2015.

\bibitem{defferrard2016convolutional}
M.~Defferrard, X.~Bresson, and P.~Vandergheynst, ``Convolutional neural
  networks on graphs with fast localized spectral filtering,'' in
  \emph{Advances in neural information processing systems}, 2016, pp.
  3844--3852.

\bibitem{shuman2013emerging}
D.~I. Shuman, S.~K. Narang, P.~Frossard, A.~Ortega, and P.~Vandergheynst, ``The
  emerging field of signal processing on graphs: Extending high-dimensional
  data analysis to networks and other irregular domains,'' \emph{IEEE signal
  processing magazine}, vol.~30, no.~3, pp. 83--98, 2013.

\bibitem{vaswani2017attention}
A.~Vaswani, N.~Shazeer, N.~Parmar, J.~Uszkoreit, L.~Jones, A.~N. Gomez,
  {\L}.~Kaiser, and I.~Polosukhin, ``Attention is all you need,'' in
  \emph{Advances in neural information processing systems}, 2017, pp.
  5998--6008.

\bibitem{sgc_wu2019simplifying}
F.~Wu, A.~H.~S. Jr., T.~Zhang, C.~Fifty, T.~Yu, and K.~Q. Weinberger,
  ``Simplifying graph convolutional networks,'' in \emph{ICML}, 2019, pp.
  6861--6871.

\bibitem{nt2019revisiting}
H.~NT and T.~Maehara, ``Revisiting graph neural networks: All we have is
  low-pass filters,'' \emph{arXiv preprint arXiv:1905.09550}, 2019.

\bibitem{pan2019learning}
S.~Pan, R.~Hu, S.-f. Fung, G.~Long, J.~Jiang, and C.~Zhang, ``Learning graph
  embedding with adversarial training methods,'' \emph{IEEE Transactions on
  Cybernetics}, vol.~50, no.~6, pp. 2475--2487, 2019.

\bibitem{sage_hamilton2017inductive}
W.~Hamilton, Z.~Ying, and J.~Leskovec, ``Inductive representation learning on
  large graphs,'' in \emph{Advances in neural information processing systems},
  2017, pp. 1024--1034.

\bibitem{clustergcn_chiang2019cluster}
W.-L. Chiang, X.~Liu, S.~Si, Y.~Li, S.~Bengio, and C.-J. Hsieh, ``Cluster-gcn:
  An efficient algorithm for training deep and large graph convolutional
  networks,'' in \emph{Proceedings of the 25th ACM SIGKDD International
  Conference on Knowledge Discovery \& Data Mining}, 2019, pp. 257--266.

\bibitem{graphsaint_Zeng2020GraphSAINT}
H.~Zeng, H.~Zhou, A.~Srivastava, R.~Kannan, and V.~Prasanna, ``Graphsaint:
  Graph sampling based inductive learning method,'' in \emph{International
  Conference on Learning Representations}, 2020.

\bibitem{diffusion_klicpera2019diffusion}
J.~Klicpera, S.~Wei{\ss}enberger, and S.~G{\"u}nnemann, ``Diffusion improves
  graph learning,'' in \emph{Advances in Neural Information Processing
  Systems}, 2019, pp. 13\,354--13\,366.

\bibitem{wu2019graph}
Z.~Wu, S.~Pan, G.~Long, J.~Jiang, and C.~Zhang, ``Graph wavenet for deep
  spatial-temporal graph modeling,'' \emph{arXiv preprint arXiv:1906.00121},
  2019.

\bibitem{wu2020connecting}
Z.~Wu, S.~Pan, G.~Long, J.~Jiang, X.~Chang, and C.~Zhang, ``Connecting the
  dots: Multivariate time series forecasting with graph neural networks,''
  \emph{arXiv preprint arXiv:2005.11650}, 2020.

\bibitem{wan2020hyperspectral}
S.~Wan, C.~Gong, P.~Zhong, S.~Pan, G.~Li, and J.~Yang, ``Hyperspectral image
  classification with context-aware dynamic graph convolutional network,''
  \emph{IEEE Transactions on Geoscience and Remote Sensing}, 2020.

\bibitem{wan2020reasoning}
G.~Wan, S.~Pan, C.~Gong, C.~Zhou, and G.~Haffari, ``Reasoning like human:
  Hierarchical reinforcement learning for knowledge graph reasoning,'' in
  \emph{International Joint Conference on Artificial Intelligence 2020}.\hskip
  1em plus 0.5em minus 0.4em\relax Association for the Advancement of
  Artificial Intelligence (AAAI), 2020, pp. 1926--1932.

\bibitem{xian2019reinforcement}
Y.~Xian, Z.~Fu, S.~Muthukrishnan, G.~De~Melo, and Y.~Zhang, ``Reinforcement
  knowledge graph reasoning for explainable recommendation,'' in
  \emph{Proceedings of the 42nd International ACM SIGIR Conference on Research
  and Development in Information Retrieval}, 2019, pp. 285--294.

\bibitem{mixedad_zhu2020mixedad}
M.~Zhu and H.~Zhu, ``Mixedad: A scalable algorithm for detecting mixed
  anomalies in attributed graphs,'' in \emph{Proceedings of the AAAI Conference
  on Artificial Intelligence}, vol.~34, no.~01, 2020, pp. 1274--1281.

\bibitem{yu2018netwalk}
W.~Yu, W.~Cheng, C.~C. Aggarwal, K.~Zhang, H.~Chen, and W.~Wang, ``Netwalk: A
  flexible deep embedding approach for anomaly detection in dynamic networks,''
  in \emph{Proceedings of the 24th ACM SIGKDD International Conference on
  Knowledge Discovery \& Data Mining}, 2018, pp. 2672--2681.

\bibitem{cl_survey_liu2020selfsupervised}
X.~Liu, F.~Zhang, Z.~Hou, Z.~Wang, L.~Mian, J.~Zhang, and J.~Tang,
  ``Self-supervised learning: Generative or contrastive,'' 2020.

\bibitem{moco_he2020momentum}
K.~He, H.~Fan, Y.~Wu, S.~Xie, and R.~Girshick, ``Momentum contrast for
  unsupervised visual representation learning,'' in \emph{Proceedings of the
  IEEE/CVF Conference on Computer Vision and Pattern Recognition}, 2020, pp.
  9729--9738.

\bibitem{grill2020bootstrap}
J.-B. Grill, F.~Strub, F.~Altché, C.~Tallec, P.~H. Richemond, E.~Buchatskaya,
  C.~Doersch, B.~A. Pires, Z.~D. Guo, M.~G. Azar, B.~Piot, K.~Kavukcuoglu,
  R.~Munos, and M.~Valko, ``Bootstrap your own latent: A new approach to
  self-supervised learning,'' 2020.

\bibitem{chen2020exploring}
X.~Chen and K.~He, ``Exploring simple siamese representation learning,'' 2020.

\bibitem{dgi_velickovic2019deep}
P.~Veličković, W.~Fedus, W.~L. Hamilton, P.~Liò, Y.~Bengio, and R.~D. Hjelm,
  ``Deep graph infomax,'' in \emph{International Conference on Learning
  Representations}, 2019.

\bibitem{mtv_hassani2020contrastive}
K.~Hassani and A.~H. Khasahmadi, ``Contrastive multi-view representation
  learning on graphs,'' in \emph{Proceedings of International Conference on
  Machine Learning}, 2020, pp. 3451--3461.

\bibitem{gcc_qiu2020gcc}
J.~Qiu, Q.~Chen, Y.~Dong, J.~Zhang, H.~Yang, M.~Ding, K.~Wang, and J.~Tang,
  ``Gcc: Graph contrastive coding for graph neural network pre-training,'' in
  \emph{Proceedings of the 26th ACM SIGKDD International Conference on
  Knowledge Discovery \& Data Mining}, 2020, pp. 1150--1160.

\bibitem{peng2020graph}
Z.~Peng, W.~Huang, M.~Luo, Q.~Zheng, Y.~Rong, T.~Xu, and J.~Huang, ``Graph
  representation learning via graphical mutual information maximization,'' in
  \emph{Proceedings of The Web Conference 2020}, 2020, pp. 259--270.

\bibitem{rwr_tong2006fast}
H.~Tong, C.~Faloutsos, and J.-Y. Pan, ``Fast random walk with restart and its
  applications,'' in \emph{Sixth international conference on data mining
  (ICDM'06)}.\hskip 1em plus 0.5em minus 0.4em\relax IEEE, 2006, pp. 613--622.

\bibitem{ff_leskovec2006sampling}
J.~Leskovec and C.~Faloutsos, ``Sampling from large graphs,'' in
  \emph{Proceedings of the 12th ACM SIGKDD international conference on
  Knowledge discovery and data mining}, 2006, pp. 631--636.

\bibitem{sen2008collective}
P.~Sen, G.~Namata, M.~Bilgic, L.~Getoor, B.~Galligher, and T.~Eliassi-Rad,
  ``Collective classification in network data,'' \emph{AI magazine}, vol.~29,
  no.~3, pp. 93--93, 2008.

\bibitem{tang2008arnetminer}
J.~Tang, J.~Zhang, L.~Yao, J.~Li, L.~Zhang, and Z.~Su, ``Arnetminer: extraction
  and mining of academic social networks,'' in \emph{Proceedings of the 14th
  ACM SIGKDD international conference on Knowledge discovery and data mining},
  2008, pp. 990--998.

\bibitem{ogb_hu2020ogb}
W.~Hu, M.~Fey, M.~Zitnik, Y.~Dong, H.~Ren, B.~Liu, M.~Catasta, and J.~Leskovec,
  ``Open graph benchmark: Datasets for machine learning on graphs,''
  \emph{arXiv preprint arXiv:2005.00687}, 2020.

\bibitem{song2007conditional}
X.~Song, M.~Wu, C.~Jermaine, and S.~Ranka, ``Conditional anomaly detection,''
  \emph{IEEE Transactions on knowledge and Data Engineering}, vol.~19, no.~5,
  pp. 631--645, 2007.

\bibitem{ding2019interactive}
K.~Ding, J.~Li, and H.~Liu, ``Interactive anomaly detection on attributed
  networks,'' in \emph{Proceedings of the Twelfth ACM International Conference
  on Web Search and Data Mining}, 2019, pp. 357--365.

\bibitem{skillicorn2007detecting}
D.~B. Skillicorn, ``Detecting anomalies in graphs,'' in \emph{2007 IEEE
  Intelligence and Security Informatics}.\hskip 1em plus 0.5em minus
  0.4em\relax IEEE, 2007, pp. 209--216.

\bibitem{adam_kingma2014adam}
D.~P. Kingma and J.~Ba, ``Adam: A method for stochastic optimization,''
  \emph{arXiv preprint arXiv:1412.6980}, 2014.

\bibitem{pca_wold1987principal}
S.~Wold, K.~Esbensen, and P.~Geladi, ``Principal component analysis,''
  \emph{Chemometrics and intelligent laboratory systems}, vol.~2, no. 1-3, pp.
  37--52, 1987.

\bibitem{paszke2019pytorch}
A.~Paszke, S.~Gross, F.~Massa, A.~Lerer, J.~Bradbury, G.~Chanan, T.~Killeen,
  Z.~Lin, N.~Gimelshein, L.~Antiga \emph{et~al.}, ``Pytorch: An imperative
  style, high-performance deep learning library,'' \emph{Advances in Neural
  Information Processing Systems}, vol.~32, pp. 8026--8037, 2019.

\bibitem{wang2019dgl}
M.~Wang, D.~Zheng, Z.~Ye, Q.~Gan, M.~Li, X.~Song, J.~Zhou, C.~Ma, L.~Yu,
  Y.~Gai, T.~Xiao, T.~He, G.~Karypis, J.~Li, and Z.~Zhang, ``Deep graph
  library: A graph-centric, highly-performant package for graph neural
  networks,'' \emph{arXiv preprint arXiv:1909.01315}, 2019.

\bibitem{pedregosa2011scikit}
F.~Pedregosa, G.~Varoquaux, A.~Gramfort, V.~Michel, B.~Thirion, O.~Grisel,
  M.~Blondel, P.~Prettenhofer, R.~Weiss, V.~Dubourg \emph{et~al.},
  ``Scikit-learn: Machine learning in python,'' \emph{the Journal of machine
  Learning research}, vol.~12, pp. 2825--2830, 2011.

\end{thebibliography}
\bibliographystyle{IEEEtran}

\balance

\end{document}